\pdfoutput=1

\documentclass[11pt,a4paper]{article}
\usepackage{acl} %
\usepackage{times}
\usepackage{latexsym}
\usepackage[T1]{fontenc}
\usepackage{bbm}
\usepackage[utf8]{inputenc}
\usepackage{graphicx}
\usepackage{url}
\usepackage{amsmath}
\usepackage{booktabs}
\usepackage{array,ragged2e}
\newcolumntype{R}[1]{>{\RaggedLeft\arraybackslash}p{#1}}
\usepackage{mdframed}
\usepackage{multirow}
\usepackage{amsfonts}
\usepackage{endnotes}
\usepackage{xspace}
\usepackage{hhline}
\usepackage{subcaption}
\usepackage{mathtools}

\usepackage{adjustbox}
\usepackage{array}
\usepackage{xcolor}
\usepackage{color, colortbl}

\newcolumntype{R}[2]{%
    >{\adjustbox{angle=#1,lap=\width-(#2)}\bgroup}%
    l%
    <{\egroup}%
}
\def\basiceval#1{\the\numexpr#1\relax}
\def\cca#1{\cellcolor{blue!#10}\ifnum #1>4\color{white!#10}\else\color{black!100}\fi{.#1}}
\def\ccb#1{\cellcolor{red!#10}\ifnum #1>4\color{white!#10}\else\color{black!100}\fi{-.#1}}
\def\ccaten#1{\cellcolor{blue!#1}\ifnum #1>40\color{white!100}\else\color{black!100}\fi{
\ifnum #1>10 .#1 \else .0#1 \fi
}}
\def\ccatentwo#1#2{\cellcolor{blue!\basiceval{#2*10}}\ifnum #2>4\color{white!#1}\else\color{black!100}\fi{#1.#2}}
\def\ccbten#1{\cellcolor{red!#1}\ifnum #1>40\color{white!#1}\else\color{black!100}\fi{
\ifnum #1>10 -.#1 \else -.0#1 \fi
}}
\def\ccbtentwo#1#2{\cellcolor{red!\basiceval{#2*5}}\ifnum #2>4\color{white!#1}\else\color{black!100}\fi{-#1.#2}}
\def\ccd#1{\cellcolor{blue!\basiceval{#1 * 2}}\ifnum #1>20\color{white!#1}\else\color{black!100}\fi{
\ifnum #1>10 .#1 \else .0#1 \fi
}}
\def\ccy#1{\cellcolor{yellow!\basiceval{#1 * 3}}\ifnum #1>50\color{white!#1}\else\color{black!100}\fi{
\ifnum #1>9 .#1 \else .0#1 \fi
}}

\DeclareMathOperator*{\argmax}{arg\,max}

\usepackage{soul}
\definecolor{darkgreen}{HTML}{38761d}
\definecolor{darkyellow}{HTML}{ffa500}
\definecolor{LimeGreen}{RGB}{179,253,148}
\definecolor{DarkLimeGreen}{RGB}{124,176,102}%
\definecolor{DarkPink}{RGB}{255,135,135}
\definecolor{OliveGreen}{rgb}{0,0.6,0}
\DeclareRobustCommand{\hlpink}[1]{{\sethlcolor{pink}\hl{#1}}}
\DeclareRobustCommand{\hlgreen}[1]{{\sethlcolor{LimeGreen}\hl{#1}}}

\definecolor{MainEvent}{RGB}{31,119,180}
\definecolor{Consequence}{RGB}{255,127,14}
\definecolor{PreviousEvent}{RGB}{44,160,44}
\definecolor{CurrentContext}{RGB}{23,190,207}
\definecolor{HistoricalEvent}{RGB}{148,103,189}
\definecolor{AnecdotalEvent}{RGB}{140,86,75}
\definecolor{Evaluation}{RGB}{227,119,194}
\definecolor{Expectation}{RGB}{188,189,34}
\colorlet{MainEvent}{MainEvent!51}
\colorlet{Consequence}{Consequence!49}
\colorlet{PreviousEvent}{PreviousEvent!51}
\colorlet{CurrentContext}{CurrentContext!53}
\colorlet{HistoricalEvent}{HistoricalEvent!51}
\colorlet{AnecdotalEvent}{AnecdotalEvent!53}
\colorlet{darkgreen}{darkgreen!70}
\colorlet{Evaluation}{Evaluation!54}
\colorlet{Expectation}{Expectation!51}
\DeclareRobustCommand{\hlMainEvent}[1]{{\sethlcolor{MainEvent}\hl{#1}}}
\DeclareRobustCommand{\hlConsequence}[1]{{\sethlcolor{Consequence}\hl{#1}}}
\DeclareRobustCommand{\hlPreviousEvent}[1]{{\sethlcolor{PreviousEvent}\hl{#1}}}
\DeclareRobustCommand{\hlCurrentContext}[1]{{\sethlcolor{CurrentContext}\hl{#1}}}
\DeclareRobustCommand{\hlHistoricalEvent}[1]{{\sethlcolor{HistoricalEvent}\hl{#1}}}
\DeclareRobustCommand{\hlAnecdotalEvent}[1]{{\sethlcolor{AnecdotalEvent}\hl{#1}}}
\DeclareRobustCommand{\hlEvaluation}[1]{{\sethlcolor{Evaluation}\hl{#1}}}
\DeclareRobustCommand{\hlExpectation}[1]{{\sethlcolor{Expectation}\hl{#1}}}

\usepackage{microtype}

\title{Sequentially Controlled Text Generation}

\author{Alexander Spangher \\ \\
  University of \\ Southern California\\ 
  \texttt{spangher@usc.edu} \\
  \\\And
  Xinyu Hua, Yao Ming \\ \\
  Bloomberg \\
  \texttt{\{xhua22, yming7\} }\\
  \texttt{@bloomberg.net} \\
  \\\And 
    Nanyun Peng \\ \\ 
  University of California\\ Los Angeles \\
  \texttt{violetpeng@cs.ucla.edu}
  }

\begin{document}
\maketitle
\begin{abstract}
While GPT-2 generates sentences that are remarkably human-like, longer documents can ramble and do not follow human-like writing structure. We study the problem of imposing structure on long-range text. We propose a novel controlled text generation task, \textit{sequentially controlled text generation}, and identify a dataset, \textit{NewsDiscourse} as a starting point for this task. We develop a sequential controlled text generation pipeline with generation and editing. We test different degrees of structural awareness and show that, in general, more structural awareness results in higher control-accuracy, grammaticality, coherency and topicality, approaching human-level writing performance. \footnote{This paper has been accepted to \textit{Findings of the 2022 Conference on Empirical Methods in Natural Language Processing.}}
\end{abstract}

\section{Introduction}

Imagine that you are tasked with: Write a ``Related Works'' section. Would it help to know the \textit{past structure} of the article (e.g. it is coming after the ``Discussion'' section)? How about the \textit{full structure} of the article (e.g. after the ``Introduction'' but before the ``Problem Statement'')?

\begin{figure*}[t]
	\centering
	\subfloat[Structure of human-written articles.]{%
		\includegraphics[width=.3\linewidth]{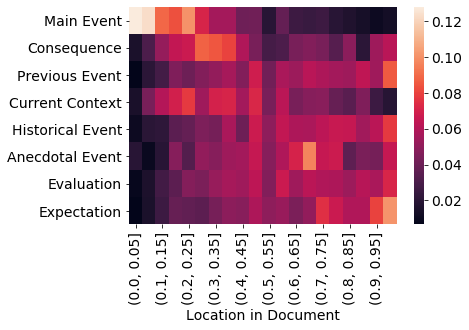}
		\label{fig:structure:human}%
	} \hspace{.1cm}
	\subfloat[Structure of naively generated GPT-2 articles]{
		\includegraphics[width=.3\linewidth]{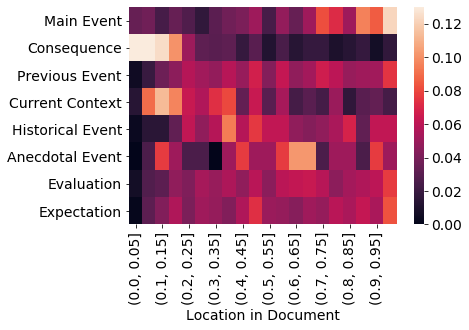}
		\label{fig:structure:naive_gpt}
	}  \hspace{.1cm}
	\subfloat[Structure of sequentially controlled GPT-2 articles.]{
		\includegraphics[width=.3\linewidth]{figures/human-written-structure.png}
		\label{fig:structure:controlled_gpt}
	}
	\caption{Discourse structure \cite{van2013news} of articles generated according to different processes. The likelihood of a tag in the $kth$ fraction of a news article is shown. Machine-generated structure is labeled by humans.}
	\label{fig:structure}
\end{figure*}

The macro-structure of text (i.e. its discourse structure \cite{po2003news}) impacts both human and machine comprehension \cite{emde2016does, sternadori2010men, lu2019attributed, zhou2020fake}. 
Although naive language models have made impressive advancements and generate fluent text \cite{radford2019language,brown2020language, beltagy2020longformer}, the text is \textit{structurally} dissimilar to human-written text (Figure \ref{fig:structure}, Section \ref{sct:results})%
. Even the well-known Ovid's Unicorn generation, which seems like a natural news article, exhibits unnatural structure (see Appendix \ref{app:ovids_unicorn}).

On the other hand, although numerous works have focused on content planning using keywords \cite{yao2019plan}, plot-design \cite{rashkin2020plotmachines} and entity tracking \cite{peng2021guiding}, macro-structural control has been relatively understudied.

\begin{figure}[t]
    \centering
    \includegraphics[width=\linewidth]{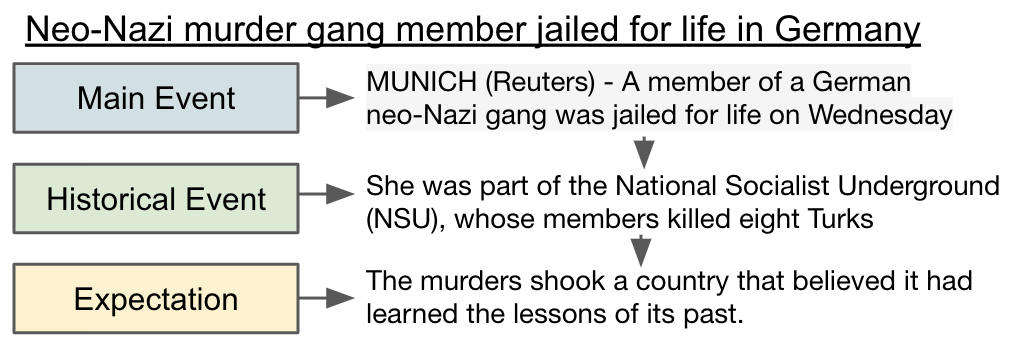}
    \caption{We study the task of \textit{sequentially-controlled generation}: generating documents exhibiting structure given by a sequence of local control codes. Shown is a news article with it's Van Dijk structure \cite{van2013news} and headline. Our models take as input the headline and discourse tags and generate a sequence of sentences. We explore the degree of structural awareness (local, past-aware or full-sequence) for controlling each sentence in the document, with the goal of generating the most structurally faithful, coherent and topical text.}
    \label{fig:cover}
\end{figure}

So, in this work, we study \textit{(1) how to impose macro-structural control on narrative text generation} and \textit{(2) how much structural awareness during generation contributes to well-structured and fluent text.} We propose a novel task, \textit{sequentially controlled text generation}. In this task, the user provides a prompt as well as a sequence of local control codes, each of which guides the generation of a single sentence. (In our experiments, we use headlines as prompts and \newcite{van2013news} discourse tags as control codes (Figure \ref{fig:cover}).)

We develop methods to address this task, expanding prior work focused on \textit{single control code generation} \cite{keskar2019ctrl, dathathri2019plug, yang2021fudge}. As in prior work, the controlled generation problem is decomposed into a \textit{discriminator} and a \textit{generator}. However, in this work, the discriminator learns to incorporate an entire \textit{sequence} of control codes. We hypothesize that information about structural intention can positively impact generative output (intuition for this is given in the hypothetical at beginning of this introduction). We show that our methods improve structural cohesion \textit{and} certain aspects of coherence over naive GPT-2 output.

Next, we hypothesize that \textit{more} structural awareness improves generation. Again, we refer to the introduction hypothetical: humans craft text according to how it fits into a document's full structure \cite{chenlo2014rhetorical}, so a generative model should similarly benefit from having such information. We test this hypothesis by varying the discriminator's conditional independence assumptions. We experiment with three different degrees of control: \textbf{local-only} (where the discriminator is only aware of the current sentences' control code), \textbf{past-aware} (where the discriminator is aware of the current sentences' control code \textit{and} all previous control codes), and \textbf{full-sequence} (where the discriminator is aware of the entire document's sequence of control codes). We show that more structural awareness, especially of past structure, helps generate the highest-quality text. Finally, we show how to re-introduce a degree of local control by combining structurally-aware generation methods with a local sentence-level editing technique. %

In summary, our novel contributions are:

\begin{itemize}
  \item We propose a novel task, \textit{sequentially controlled text generation} and identify a discourse schema \cite{van2013news} and dataset \cite{choubey-etal-2020-discourse} to explore this task (Sections \ref{sct:problem_statement}, \ref{sct:datasets}).
  \item We combine two different approaches in controlled text generation: \textit{generation} and \textit{editing}, and show that the highest-quality text is generated when both of these approaches are used (Section \ref{sct:methodology}).
  \item We use our methods to study the \textit{degree} of structural control that yields the highest-quality text: \textit{local}, \textit{past-aware} and \textit{full-sequence} control. We show that overall, \textit{full-sequence} produces optimal text over an array of metrics (Section \ref{sct:results}).
\end{itemize}

We see this work opening the door to a variety of follow-on directions: giving users control over the macro-structure of their generated output can allow users to quickly prototype different structures for their work. It can allow them to work in tandem with a generative algorithm to infill missing structural components in a piece of writing\footnote{Perhaps aiding in human-in-the-loop computational journalism \cite{cohen2011computational}}. It might even allow them produce different versions of the same story for readers with different reading preferences. Finally, we also see macro-structural control providing a natural complement to, and being used in tandem with, other forms of controlled generation, like fact-aware generation \cite{logan2019barack} or creative generation \cite{goldfarb2020content, tian2022zero, peng2022controllable} to yield more engaging and useful generative content. 

\section{Problem Statement}
\label{sct:problem_statement}

We assume, as input, a headline sentence, $X_0$, and a sequence of control codes $\vec{c} = c_1,...,c_S$ of length $S$ (i.e., one for each sentence we wish to generate in the document. \textit{Adjacent codes can be of the same type}.) We wish to produce, as output, a document $\mathbf{X}$ of length $S$ as a sequence of sentences $\mathbf{X}=X_1,...,X_S$, each composed of a sequence of words $X_k=x_1,...,x_{n_k}$ of length $n_k$. 

We define the sequentially controlled text generation objective as:
\begin{multline}
\label{eq:sequentially_controlled_gen}
	p(x| \vec{c}) = \prod_{k=1}^S\prod_{i=1}^{n_k} \underbrace{p(x_i | x_{<i}, X_{<k}, \vec{c})}_{t_1: \text{ word likelihood}}
\end{multline}

where $x_i$ is a word in sentence $k$, $x_{<i}$ are the preceding words, $X_{<k}$ are the preceding sentences (including the headline, $X_0$). $c_k$ is the control code for $k$. We assume that $\vec{c}$, the entire sequence of control-codes for a document, is given.

We use Bayes rule to factorize $t_1$ into:
\begin{multline}
	\label{eq:factorized_sequential_control}
	\propto \underbrace{ p(x_i| x_{<i}, X_{<k}) }_{t_2: \text{ naive word likelihood}} \underbrace{p(\vec{c} | x_i, x_{<i}, X_{<k})}_{t_3:\text{ class likelihood}}
\end{multline}

$t_2$ is calculated using a standard pretrained language model (PTLM) and $t_3$ is calculated by a trained discriminator. This allows us to maximally re-use naively trained language models and, as we show, is far more resource efficient than fine-tuning a prompt-based model. %

Three approximations for $t_3$ are:
\begin{multline}
	\label{eq:local_only}
	\text{\textbf{Local-Only}}\quad\quad\quad
	t_3 \approx p(c_s | x_i, x_{<i}, X_{<s}) 
\end{multline}
In the local-only model, we assume each control code $c_k$ is conditionally independent of other control codes given $x_i$. Thus, our generator model $t_1$ is made aware only of local structure: the control code $c_k$ pertaining to the current sentence, $k$. Because of this conditional independence assumption, \textit{local-only} control is similar to prior work that used only single-control codes, where the goal was to generate a single sentence $p(x|c) = \prod_{i=1}^n p(x_i | c)$ \cite{keskar2019ctrl}. However, we show that we can remove these independence assumptions and study more complicated structural control which, as we show later, produces more coherent output.
\begin{multline}
	\label{eq:past_aware}
	\text{\textbf{Past-Aware}}
	\\ t_3 \approx \prod_{j=1}^{k} p(c_{j} |x_i, x_{<i}, X_{<k}, c_{<j}) 
\end{multline}

In the past-aware model, we assume autoregressive dependence between control codes, conditioned on $x$. Control codes for future sentences, $c_{>k}$, are conditionally independent. In Equation \ref{eq:sequentially_controlled_gen}, this results in $x_i$ being dependent on $c_{k}$ and the sequence of control codes, $c_{<k}$.
\begin{multline}
	\label{eq:full_control}
	\text{\textbf{Full-Sequence}}
	\\ t_3 = \prod_{j=1}^{S} p(c_{j} |x_i, x_{<i}, X_{<k}, c_{<j}) 
\end{multline}

In the full-sequence model, we make no conditional independence assumptions. %

We can restrict both the past-aware and the full-sequence approximations to a sliding window around sentence $s$\footnote{i.e. $t_3$ ranges only from $j=k-w...k+w$ instead of the full sequence of sentences. In practice, we use $w=3$.}. We can also add a prior on $p(\vec{c})$ to induce a discount factor\footnote{The form of our prior is: $t_3 = \prod_{j=1}^{S} m(i, j) p(c_{j} |x_i, x_{<i}, X_{<k}, c_{<j})$, where $m(i, j) = b^{|i-j|}$. We experiment with $b=[.33, .66, 1]$.}. This focuses the generator on control code $c_k$ and down-weights surrounding control codes.

In the next section, we show how to model these objectives. %
We first describe the discriminator we use as our control-code model, the controlled generation techniques and the editing techniques we adapt. 

\begin{figure*}
	\centering
	\includegraphics[width=.75\linewidth]{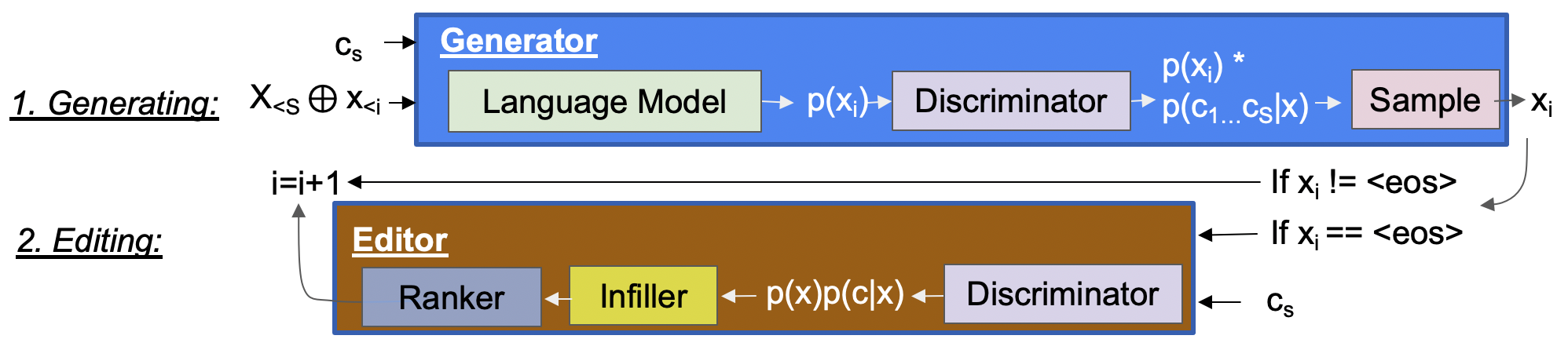}
	\caption{\textbf{Generation process}. First, we perturb the output of a language model using a structurally-aware classifier to approximate $p(x_i|x_{<i}, X_{<k})p(\vec{c} | x_{<i}, X_{<k})$ and generate word $x_i$ by sampling from the perturbed distribution . When we generate an $<eos>$ token, we edit the sentence. We use a discriminator to identify class-salient words to mask, generating masked sentence $M$, and infill to boost class likelihood.}
	\label{fig:gen_flow}
\end{figure*}

\section{Methodology}
\label{sct:methodology}

As described in Section \ref{sct:problem_statement}, we can efficiently do generation by combining a \textit{naively-trained} language model with a discriminator. \textit{Hence, the discriminator is the main architectural component that allows us to incorporate inter-dependencies between control code sequences.} We start by describing how our discriminator models different degrees of structural awareness (Equations \ref{eq:local_only}, \ref{eq:past_aware} and \ref{eq:full_control}) in Section \ref{sct:methodology:discriminator}. 

We design a generation pipeline to balance \textit{structural} and \textit{local awareness}. The flow we use to accomplish this is depicted in Figure \ref{fig:gen_flow}. The first step is \textbf{Generation}. Here, we sample each word, $x_i$ using techniques described in Section \ref{sct:methodology:gen_methods} which allow us to leverage our discriminator to impose \textit{structural control}. %
When we have completed a sentence, we move to \textbf{Editing}. Here, we edit the sentence to further impose \textit{local control} on each sentence, updating $x$ to optimize a variation of Equation \ref{eq:sequentially_controlled_gen}: $p(x_i | x_{-i}, c_k)$, discussed in Section \ref{sct:methodology:editing_method}. 

\subsection{Discriminator}
\label{sct:methodology:discriminator}

The discriminator we construct takes as input a sequence of sentences ($\mathbf{X}$) and a sequence of local control tags ($\vec{c}$). Our architecture combines a sentence-classification model, similar to that used in \cite{spangher2021multitask}, with a separate label embedding architecture to incorporate knowledge of $c_{<j}$. Hence, we can make predictions for $c_j$ based not only on $x$, but prior tags, $c_{<j}$, allowing us to model structural dependencies (Equation \ref{eq:factorized_sequential_control}). For a full description of architecture, see Appendix \ref{app:implementation_details}.

We train it to model local-only, past-aware and full-sequence control variants expressed in Section \ref{sct:problem_statement}: we train separate prediction heads  %
to make predictions on $c_{k-w}, ... c_k, ... c_{k+w}$, i.e. labels from $-w,...,+w$ steps away from current sentence $k$\footnote{Note: we still factor label-sequences autoregressively, as in  Equations \ref{eq:past_aware} and \ref{eq:full_control} and learn each prediction head separately. However, keeping separate heads allows the model more flexibility in predicting how attributes of a sentence might predict future or past tags. Preliminary experiments show that this approach outperforms learning a single head for all labels.}. %
For local-only control (Equation \ref{eq:local_only}) we only use predicted probabilities from the main head, $k$. In past-aware control (Equation \ref{eq:past_aware}), we multiply predicted probabilities from heads prior to the current sentence $<k$, and for \textbf{full-sequence} control, we multiply predicted probabilities from all heads.\footnote{For the editing operation, the discriminator is trained without the contexualizing layer (i.e. Transformer and $a_i$ layers are not used) because gradients need to be computed that pertain only to the sentence being edited, not previous sentences.} We now describe how we use these predictions. 

\subsection{Generation}
\label{sct:methodology:gen_methods}

We combine our discriminator's predictions with a naive PTLM to solve Equation \ref{eq:factorized_sequential_control} in two different ways: \textbf{Hidden-State Control}, based on \cite{dathathri2019plug} and \textbf{Direct Probability}, based on \cite{yang2021fudge}. %

\textbf{Hidden-State Control (HSC):} \newcite{wolf2019huggingface}'s GPT-2 implementation caches hidden states $H$ to produce logits approximating $p(x_i | x_{<i})$. We perturb these hidden states $H$, resulting in $\hat{H}$ that produce logits approximating Equation \ref{eq:sequentially_controlled_gen} instead. We generate $H$ from a naive PTLM and use this to make a prediction $\hat{c}$ using our discriminator%
. We then calculate the loss $L(\hat{c}, c)$ and backpropagate to $H$ to derive $\hat{H}$. %

\textbf{Direct-Probability Control (DPC):} We calculate $p(x_i | x_{<i}, X_{<s})$ to identify the 200 most likely $x_i$ under the naive language model, $|x_{i,j}|_{j=0}^{200}$. Then we calculate $p(c_s | x_{i, j}, x_{<i}, X_{<s}, c_{-s})$ for each $x_{i,j}$ using our discriminator. We directly multiply these probabilities to calculate Equation \ref{eq:sequentially_controlled_gen}\footnote{Note that DPC has the advantage of being simpler to implement and batch-parallelizable. However, the restriction to the top $k=200$ words selected according to $p(x_i | x_{<i}, X_{<s})$ means that we might be limiting discriminator perturbation of word-selection.}.

Note that the HSC and DPC algorithms are extensions of previous work: the difference is that here they are used to model control code \textit{sequences} rather than \textit{single} tags. \textit{The key components that allow this is our discriminator, which makes predictions based on label sequences, and our algorithm which, as shown in Figure \ref{fig:gen_flow}, increments codes each time an \textit{<eos>} token is generated.}

\begin{table*}
\small
\begin{tabular}{p{2cm}p{13.4cm}}
\toprule
Headline    & UPDATE 1-Belarus food retailer lines up landmark overseas IPO \\
\midrule
\hlMainEvent{Main Event} & SILVER SPRING, Colo. (AP) — A company in Finland is poised to list its first international grocery chain in the United States after a yearlong search, raising concerns about potential corporate conflicts of interest and legal issues around using foreign intellectual property.<|endoftext|> \\
\hlPreviousEvent{Previous Event} & The world’s largest retailer of meat and fish products hopes to become one day the leader in convenience groceries.<|endoftext|> \\
\hlExpectation{Expectation} & It says it could become a leader in fresh fruits and vegetables, as well.<|endoftext|> \\
\hlEvaluation{Evaluation} & About 300 Finnish companies have applied to list on the New York Stock Exchange under a new plan to bypass the U. S. government and create ”licenses” that allow them in many other countries, according to a regulatory filing.<|endoftext|> \\
\hlEvaluation{Evaluation} & The European Union prohibits using foreign intellectual property for purposes outside the country, and Finland’s National Stock Exchange does not require approval before selling a company’lls share in its new company.<|endoftext|> \\
\hlEvaluation{Evaluation} & "What’s at stake here is transparency, a sense of fairness to all the stakeholders.<|endoftext|> \\
\hlEvaluation{Evaluation} & I don’t think it’s right for companies to have intellectual property rights," says Michael Vakilainen, an independent analyst who has tracked the company since 2008.<|endoftext|> \\
\hlExpectation{Expectation} & He says there are potential conflicts of interest, because one partner is the government.<|endoftext|> \\
\hlExpectation{Expectation} & "What if you’re a government contractor?"<|endoftext|> \\
\bottomrule
\end{tabular}
\caption{Sample document generated. Generation Method = Direct Prob. Control. Structure = Past Aware. Edited = False. (Hyperparams = $\gamma = .75$, $b = .33$)}
\end{table*}

\begin{figure}[t]
    \centering
    \includegraphics[width=.7\linewidth]{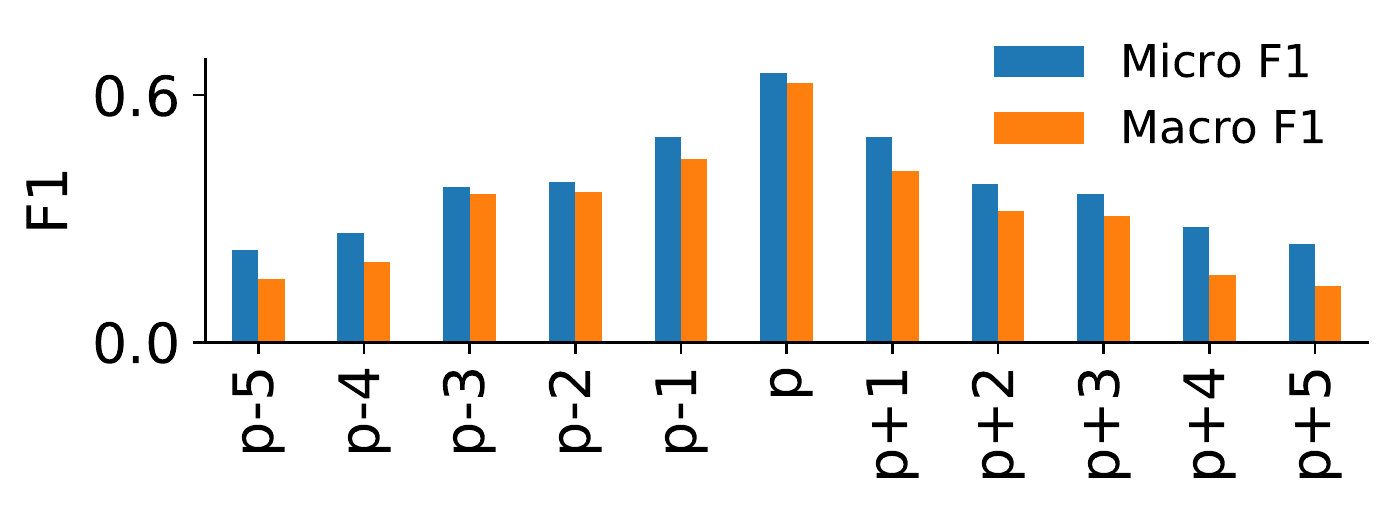}
    \caption{\textbf{Discriminator performance} on test data. F1 scores for $p(c_j | X_{<k}, x_{<i}, c_{<j})$ predictions. Sentence index $k$ and word index $i$ are fixed: we show error for using the current sentence to predict all past, current and future labels. }
    \label{fig:nn_model:f1_scores}
\end{figure}

\subsection{Editing}
\label{sct:methodology:editing_method}

After we have finished generating a sentence, we edit it to introduce more discourse markers of the local control code.

We identify words in our input sequence that have the most impact on control-code prediction by using the gradient on our input sentence of the discriminator's loss onto tokens and masking full words, following \newcite{ross-etal-2021-explaining}. We use only the current sentence prediction made by our discriminator (i.e. Equation \ref{eq:local_only}), so that we impose local control on the sequence even in settings where the generator imposes structural control.

We cull the high-gradient words based on heuristics\footnote{Words that are \textit{not} proper nouns, named entities (except the DATE class) or adjectives, as we find these categories are more likely to be topic words spuriously correlated with control-codes.} to encourage the editor to introduce explicit discourse markers. We fine-tune a label-aware infilling model \cite{raffel2019exploring} to generate candidate edits\footnote{A T5 model trained using a specific input template incorporating the label. E.g. \texttt{label: Background. text: The senator <MASK> to the courtroom to <MASK>.}} given the masked input. We mask and infill until we have generated a sentence that has an increased likelihood $p(c_k | \hat{x_k}) > p(c_k | x_k)$, and generate edit candidates ($n=10$). We select edits on the basis of class likelihood and perplexity\footnote{Perplexity of the entire generated document so far is used as a selection criteria, $PPL(x_k \oplus X_{<k})$, to encourage edits preserving the logical flow of the document.}. 

For more comparison and distinction from previous work for both Generation and Editing, see Appendix \ref{app:editing:comparison}, \ref{app:method_comparison}. %

\section{Datasets and Schema}
\label{sct:datasets}

The form of sequential control we study is \textit{discourse}: i.e. the functional role sentences play in a document's larger argumentative purpose. We use a news discourse schema proposed by \newcite{van2013news}. \newcite{choubey-etal-2020-discourse} apply this schema and annotate a dataset, \textit{NewsDiscourse}, consisting of 802 articles from 3 outlets\footnote{nytimes.com, reuters.com and xinhuanet.com}, tagged on the sentence level. %
Their schema consists of 9 classes: $\{$ \hlMainEvent{Main Event}, \hlConsequence{Consequence}, \hlCurrentContext{Current Context}, \hlPreviousEvent{Previous Event}, \hlHistoricalEvent{Historical Event}, \hlAnecdotalEvent{Anecdotal Event}, \hlEvaluation{Evaluation}, \hlExpectation{Expectation} $\}$.\footnote{For a detailed class description, see Appendix \ref{app:van_dijk_schema}}. \textit{Although each sentence is tagged with a code, codes often repeat}. For example, an entire paragraph can be tagged with \hlMainEvent{Main Event} sentences. We show a partial sample in Figure \ref{fig:cover}. We adopt this schema to describe each news article's structure.%

We also use a dataset of unlabeled news articles\footnote{\url{kaggle.com/snapcrack/all-the-news}. Dataset originally collected from \url{archive.org}. We filter to articles from nytimes.com and reuters.com.} to fine-tune a GPT-2 model for news. We sample 30,000 documents from this dataset in a manner so that the distribution of sentence-lengths matches the distribution of sentence lengths in the \newcite{choubey-etal-2020-discourse} dataset.

\section{Implementation Details}

We fine-tune a \texttt{GPT2-base} model on a large news corpus with a max word-piece length=$2048$\footnote{Rather than $1024$ in \cite{radford2019language}. We observe that $>99\%$ of human-generated news articles were shorter than $2048$ word pieces.}. We use this to generate naive PTLM language-modeling \textit{as well as} sentence-embeddings in our Discrimination model. Further implementation details are discussed in Appendix \ref{app:implementation_details}.%

We discuss the discriminator results here briefly. As shown in Figure \ref{fig:nn_model:f1_scores}, the primary head, $p$, has a Micro F1-score of $.65$, which approaches state-of-the-art on this dataset\footnote{$.71$ Micro-F1 in \newcite{spangher2021multitask}, which used auxiliary datasets.}. However, performance degrades rapidly for heads farther from $p$. For more results on discriminator performance, including experimental variations, see Appendix \ref{app:implementation_details:discriminator}. %

\section{Experiments}
\label{sct:exp}

We sample 10 documents from the test set of our discourse dataset ($n=200$) to test different pipeline settings. The input to our models is \textbf{a headline (as a prompt) and the \textit{full sequence of gold-truth discourse labels}} of that document. 

\paragraph{Baselines} We compare our experimental pipelines (Section \ref{sct:methodology}) with the following baselines: (1) \textbf{Naive GPT-2} generation given only the headline as input (i.e. no control codes), (2) a fine-tuned \textbf{Prompting} approach and (3) the original \textbf{Human}-written articles. 

For (2), we directly train a class-conditional language model to generate text by including labels in the prompt, as in \cite{keskar2019ctrl}. Local-only prompting is achieved by only including the local control code (and prior generated sentences) in the prompt, and updating the prompt to generate a new sentence. For past-aware prompting, we include all control codes prior to our current sentence in the prompt, and update on every new sentence. Finally, for full-sequence prompting, we including the full sequence of control codes in the prompt. (See Appendix \ref{app:baseline_2} for more details and examples of prompt design.) 

For each of these baselines, we test with and without editing (with the human-written text being edited by our algorithm in \textbf{Human} and with the generated text in all other trials being edited).

\paragraph{Evaluation} For all pipelines, we select the best hyperparameter configurations based on perplexity and model-assigned class likelihood. Then, we manually annotate each generated document for 4 metrics: Accuracy (0-1)\footnote{Accuracy: how close a generated sentence matches the discourse function of the gold-truth label for that sentence.} Grammar (1-5)\footnote{Grammar: how grammatical \textit{and} locally coherent a sentence is}, Logical Flow (1-5)\footnote{Logical Flow: how well a sentence functions in the flow of the story} and Topicality (1-5)\footnote{How well each sentence corresponds to the original headline of the article.}. We recruit two expert annotators with journalism experience to perform annotations blindly without awareness to which generation pipeline was used, and find moderate agreement $\kappa \in [.36, .55]$ across all categories. For more details, see Appendix \ref{app:annotation}.  We record model-dependent and non-model automatic metrics used by \newcite{see2019massively}, described further in Appendix \ref{app:metrics}.

\section{Results}
\label{sct:results}

\paragraph{Best Overall Trial} We show automatic and human metrics for the subset of pipelines with top-performing hyperparameters in Table \ref{tab:results}. In general, the highest-performing generation pipelines are all variations of DPC with either past-aware, or full-sequence structural control. 

We observe that DPC with past-aware control and editing has the highest class-label accuracy, nearly approaching the human trials. The top performing pipelines for grammar and topicality are DPC with full-Sequence control and without editing. GPT-2 performed best only for Logical Flow, which was surprising but could perhaps be because the unconstrained nature of GPT-2's generation allowed it to hallucinate a flow that seemed consistent even if it was poorly structured.

\paragraph{Effect of Different Pipeline Components} We show the distributional shifts in performance across all trials, in Figures \ref{fig:boxplots:discrim}, \ref{fig:boxplots:edited}\footnote{And \ref{fig:boxplots:generation}, in Appendix \ref{app:method_comparison}.}. Structural control has a largely positive effect on generated text. In Figure \ref{fig:boxplots:discrim}, we find that Full-Sequence models are, on average, able to generate the most label-accurate sentences with the best grammar, logical flow and topicality.  Finally, editing improves accuracy, grammar and logical flow (Figure \ref{fig:boxplots:edited}.)

The original human-generated text is our gold-standard, and it is highly class-accurate, grammatical, coherent and topical. Interestingly, as seen in Table \ref{tab:results}, editing can \textit{also} be applied to human-written text to boost label accuracy, but at the expense of coherence.

\begin{table*}[t]
	\small
    \begin{tabular}{p{1.4cm}p{1.2cm}p{1.2cm}p{1.1cm}p{1.1cm}p{1.2cm}p{1.3cm}p{1.1cm}p{1.1cm}p{1cm}}
	\toprule 
	&&
	\multicolumn{4}{c}{Human-Annotated Metrics} & 
	\multicolumn{4}{c}{Automatic Metrics} \\
	\cmidrule(lr){3-6}
	\cmidrule(lr){7-10}
	Gener-ation & Struct-ure  &  Label Acc. $\uparrow$ (0-100) &  Gram-mar $\uparrow$  (1-5) &  Logical Flow $\uparrow$ (1-5) &  On-Topic $\uparrow$ (1-5) &      Perplex. $\downarrow$ & Diverse Ngrams $\uparrow$ (\%) &  Sent. Len.** & Unseen Words $\downarrow$ (\%) \\
    \cmidrule(lr){1-2}
    \cmidrule(lr){3-6}
	\cmidrule(lr){7-10}
    \textbf{Naive GPT-2} &     &  20.0/64.4  &  4.2/4.5        &    4.7/4.3           &         4.6/4.2 &     48.2/45.4 &             7.1/8.3 &   24.9/\textbf{38.8} &          4.7/3.2 \\
    \cmidrule(lr){1-2}
    \cmidrule(lr){3-6}
	\cmidrule(lr){7-10}
    \multirow{3}{1.4cm}{\textbf{Gen-Base:} Prompt} & local & 
                                  22.2/51.1 &         2.8/3.9 &              2.4/3.0 &         2.3/2.8 &     24.4/43.4 &           3.7/6.5 &   39.7/32.4 &         10.6/8.7 \\
                        & past &  20.0/31.1 &         2.9/3.6 &              2.4/2.9 &         2.3/3.7 &     52.2/32.0 &           5.0/4.5 &   35.0/44.5 &          9.3/7.1 \\
                        & full &  46.7/64.4 &         4.4/4.4 &              3.6/3.7 &         3.9/3.5 &     42.5/49.2 &           7.3/7.8 &   35.5/42.6 &          4.6/4.9 \\
    \cmidrule(lr){1-2}
    \cmidrule(lr){3-6}
	\cmidrule(lr){7-10}
    \multirow{3}{1.4cm}{\textbf{Method \#1:} HSC} & local &  
                                  28.9/42.2 &        3.3/3.7 &               2.7/3.2 &        3.1/3.4 &    246.4/115.5 &           7.0/6.9 &   16.2/17.5 &         8.0/6.9 \\
                        & past &  44.4/60.0 &        3.4/3.8 &               3.0/3.0 &        3.2/3.3 &    178.3/147.4 &           7.5/7.5 &   14.8/18.8 &         8.1/6.7 \\
                        & full &  55.6/68.9 &        3.5/4.2 &               4.0/3.7 &        4.2/4.3 &    134.5/129.6 &           7.2/7.8 &   17.3/20.7 &         7.0/7.1 \\
    \cmidrule(lr){1-2}
    \cmidrule(lr){3-6}
	\cmidrule(lr){7-10}
    \multirow{3}{1.4cm}{\textbf{Method \#2:} DPC} & local &  
                                  44.4/64.4 &        4.0/4.4 &               3.6/4.1 &        3.8/3.5 &      42.1/39.9 &           5.8/8.3 &     24.8/42.6 &        4.7/3.0 \\
                        & past &  64.4/\textbf{88.9} &        4.5/4.6 &               \textbf{4.4}/4.3 &        4.4/4.5 &     \textbf{37.0}/42.2 &           7.9/\textbf{8.4} &     33.1/42.7 &        3.9/\textbf{3.1} \\
                        & full &  66.7/68.9 &        \textbf{4.7}/4.5 &               4.3/4.3 &       \textbf{4.7}/4.4 &      42.3/45.6 &           8.0/8.1 &     28.2/40.4 &        4.3/3.3 \\
    \cmidrule(lr){1-2}
    \cmidrule(lr){3-6}
	\cmidrule(lr){7-10}
    \textbf{Human} &           &  93.3/\textbf{95.6} &        \textbf{4.9}/4.7 &               \textbf{4.9}/4.7 &        \textbf{4.9/4.9} &      \textbf{34.2}/41.0 &           \textbf{8.7/8.7} &     \textbf{37.9}/39.6 &        4.2/4.5 \\
\bottomrule
\end{tabular}
\caption{Metrics on different trial runs. Each cell shows Unedited/Edited variants. (Hyperparams = $\gamma = .75$, $b = .33$). ** Optimal sentence length is determined relative human generation, i.e. $\min |x - 37.9|$.}
\label{tab:results}
\end{table*}

\begin{figure}[t]
     \centering
     \subfloat[]{
        \includegraphics[width=.48\linewidth]{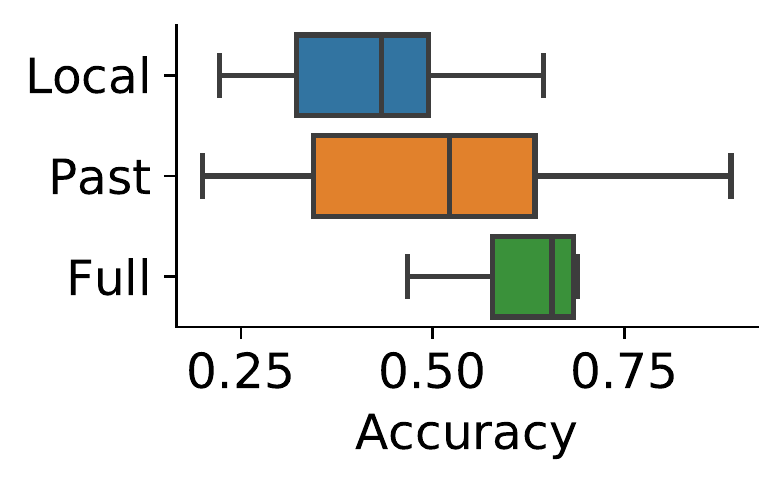}
        \label{fig:boxplots:discrim:acc}
     }
    \subfloat[]{
        \includegraphics[width=.48\linewidth]{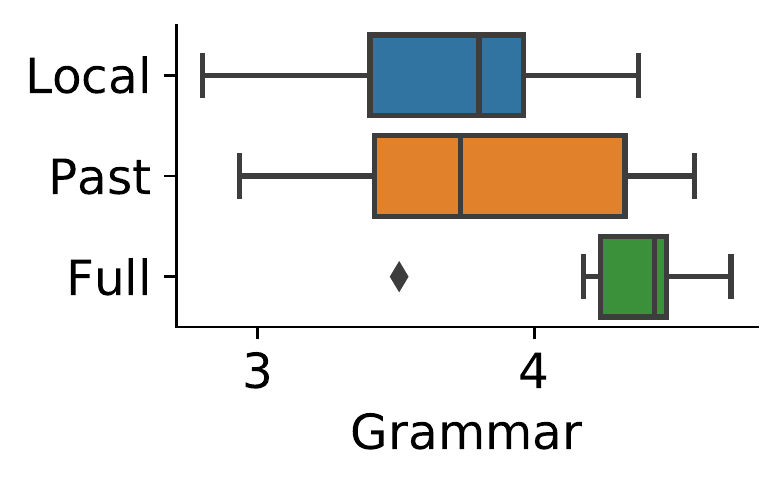}
        \label{fig:boxplots:discrim:grammar}
    } \\ 
    \subfloat[]{
        \includegraphics[width=.48\linewidth]{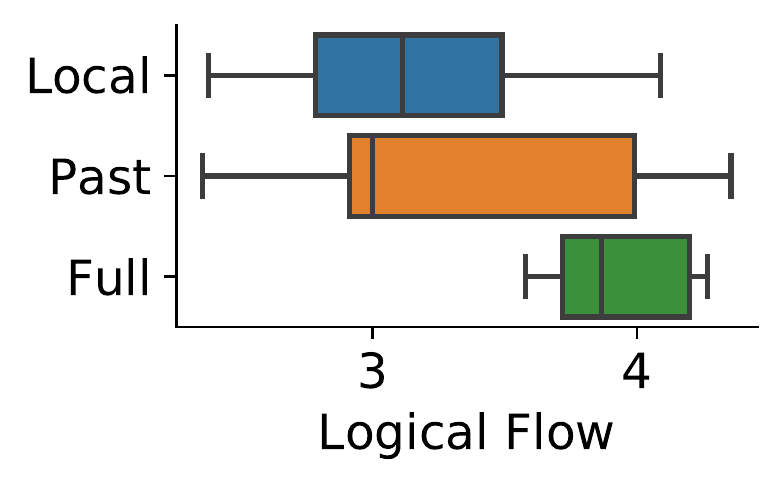}
        \label{fig:boxplots:discrim:coherence}
    }
    \subfloat[]{
        \includegraphics[width=.48\linewidth]{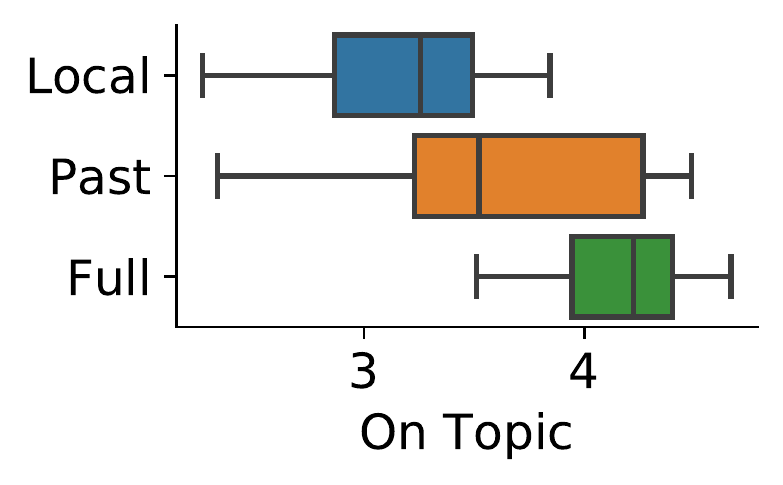}
        \label{fig:boxplots:discrim:topical}
    }
    \caption{Comparison of different structural control methods across different pipelines and hyperparameters.}
    \label{fig:boxplots:discrim}
\end{figure}

\begin{figure}[t]
     \centering
     \subfloat[]{
        \includegraphics[width=.48\linewidth]{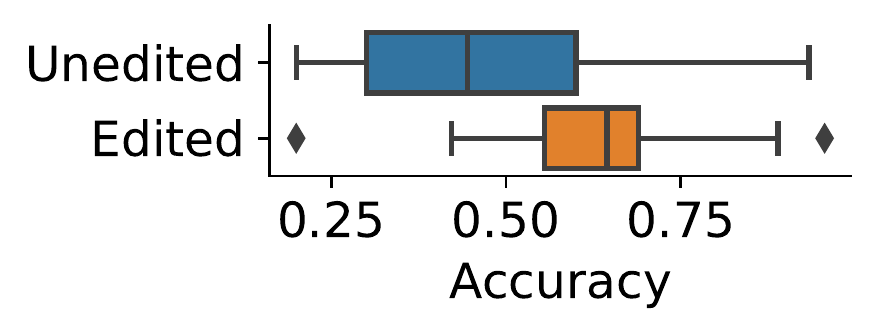}
        \label{fig:boxplots:edited:acc}
     }
    \subfloat[]{
        \includegraphics[width=.48\linewidth]{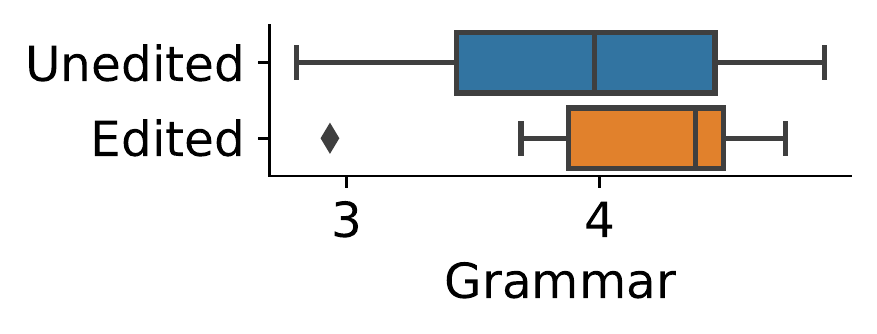}
        \label{fig:boxplots:edited:grammar}
    } \\ 
    \subfloat[]{
        \includegraphics[width=.48\linewidth]{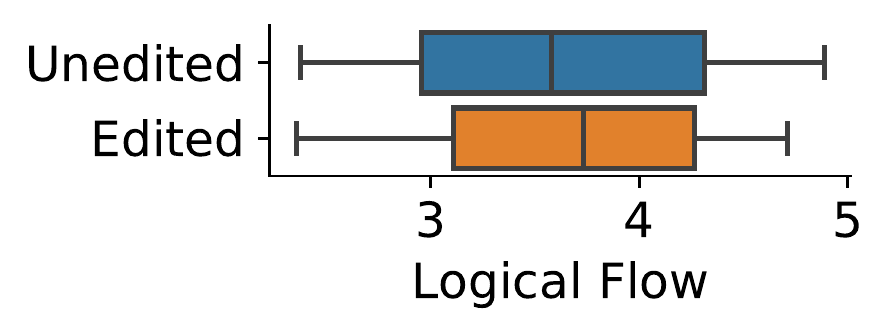}
        \label{fig:boxplots:edited:coherence}
    }
    \subfloat[]{
        \includegraphics[width=.48\linewidth]{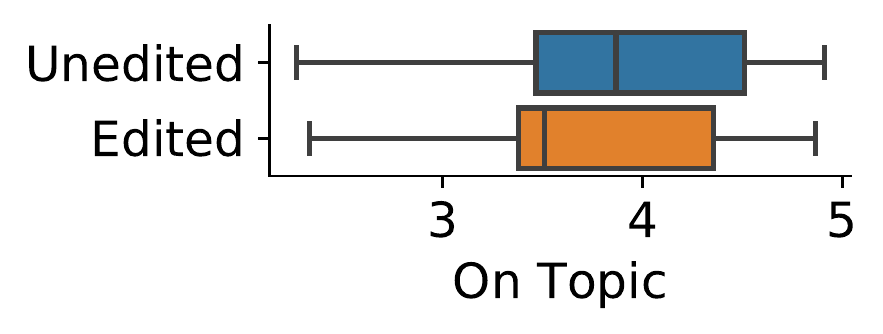}
        \label{fig:boxplots:edited:topical}
    }
    \caption{The effect of editing, across different pipelines and hyperparameters.}
    \label{fig:boxplots:edited}
\end{figure}

\section{Discussion}
\label{sct:discussion}

We set out to answer two questions in this research: (1) whether we could impose structural control over generated documents and (2) what kinds of structural control (local-only, past-aware, or full-sequence) had the greatest effect on discourse, flow, topicality and grammaticality. Our novel pipelines, which extend various discriminator-based approaches for generation and editing, approach human-level performance. However, a gap between our model's output and human-generated text still remains across all metrics, suggesting the need for more research.

\paragraph{Insight \#1: Some structural information improves all metrics of quality.} Our structural exploration suggests that, for the best-performing pipelines, \textit{past} structural information (along with editing) boosts class accuracy the most, but knowledge of the full-sequence does not. In the analogy given in the Introduction, this equates to: to write a ``Related Works'' section, it helps to know that it comes after the ``Introduction'' vs. the ``Discussion'', but not information of what sections come after. This is perhaps because enough signal is already given by the past sequence and the full sequence just adds more noise. However, full-sequence information does yield the best grammar and topicality. This might indicate a regularizing role played by the full-sequence. In general, we suspect that past-aware modeling and editing both push the model more towards the class label at the expense of topicality, flow and grammar, while full-sequence does the opposite. In practice, some combination of these pipeline components might be desired.

\paragraph{Insight \#2: Weak discriminators can still impose accurate control.} At $.61$ macro F1, our discriminator is a relatively weak classifier. Previous work in classifier-based controlled text generation used large training datasets and classifiers that routinely scored above $.8$ F1 \cite{dathathri2019plug, yang2021fudge}. The weakness of our discriminator is one reason why HSC may have performed poorly. However, in other trials we see strong accuracy. 
Thus, even with a weak classifier, we can control generation. This might be because even a weak discriminator can still give relative differences between generation that does or does match the control code.

\paragraph{Insight \#3: Evaluating text candidates using multiple model's perplexity might result in better selections.} Just as surprisingly, editing also has an overall average positive effect on generation accuracy \textit{and} generation \textit{quality} (Figure \ref{fig:boxplots:edited}). We had hypothesized that, because the editor makes locally-aware infilling decisions, it would improve class-accuracy but hurt other metrics of document quality, like topicality and flow. Indeed, for the top-performing trials, like DPC and Human, Editing only improves class accuracy. However, grammar and flow improves in other trials. This could be because, as mentioned in Section \ref{sct:methodology:editing_method}, we selected candidates based on how well they make sense in the document. %
This also suggests that using multiple PTLMs to select for better quality combines different virtues of each model.

\paragraph{Error Analysis:} We observed that sentence tokenizing remained a huge challenge. Many of the grammar errors that our annotators observed were from sentences that ended early, i.e. after decimal points. Indeed, the correlation between sentence-length and grammar is relatively high ($r=.34$). One reason for this could be that error-prone sentence tokenizing models provided faulty training data during pretrainining of LMs. This will continue to hinder document-level structural work, which often relies on a model accurately ending a sentence. Another observation, in Table \ref{tab:results}, is that perplexity doesn't necessarily correlate with human judgements of quality, especially for more complex writing like \textit{Financial} news reporting. %

\section{Related Work}

\paragraph{Discourse-Aware Narrative Text Generation.} 
Generating narrative text, such as news articles and scientific reports, has been a long standing problem in NLP. Early work relies on template~\cite{xu-etal-2018-skeleton,wiseman-etal-2018-learning}, rules~\cite{ahn-etal-2016-improving,leppanen-toivonen-2021-baseline}, or specialized architectures ~\cite{fan-etal-2018-hierarchical,bosselut-etal-2018-discourse} that are hard to generalize.

Recently, pre-trained Transformers have shown impressive capabilities to produce fluent text, and there are few works that seek to adapt them to document-level generation with appropriate discourse structures. One work, DiscoDVT \cite{ji2021discodvt}, uses a discrete variational auto-encoder (VAE) with a latent space guided by explicit Penn Discourse Treebank (PDTB) relations \cite{prasad2008penn}. We are excited by this work, which shows strong improvements in coherence. While our work is able to learn from more abstract structural tags rather than low-level PDTB relations, our approach is fully supervised. We are excited by the semi-supervised nature of \newcite{ji2021discodvt}'s approach, which may allow it to learn discourse structures that are less well annotated. Also, possible extensions of this approach using hierarchical discrete VAEs \cite{razavi2019generating} or diffusion models \cite{li2022diffusion} might provide users' control over higher-order macro-structures in text, such as those explored in \cite{spangher2021don}.

\paragraph{Controlled Generation}
The black-box nature of neural generation models poses challenges for many real-world applications~\cite{wiseman-etal-2017-challenges,holtzman2019curious}. Researchers have designed various techniques to control the syntactic structure~\cite{goyal-durrett-2020-neural}, sentiment~\cite{hu2017toward,fu2018style, luo-etal-2019-learning}, and language style~\cite{niu-bansal-2018-polite,cao-wang-2021-inference}.  
Most notably, the CTRL model~\cite{keskar2019ctrl} conditions the output by incorporating textual control codes during the pre-training stage. However, such training is resource-intensive and requires large datasets. Alternatively, PPLM~\cite{dathathri2019plug}, FUDGE~\cite{yang2021fudge}, GeDI~\cite{krause-etal-2021-gedi-generative}, and NADO~\cite{meng2022nado} achieve inference-time control through either directly manipulating the generator's hidden states, or adjusting the probabilistic distribution over the output vocabulary. Our work differs from prior work in that we tackle structured control instead of a single attribute. Our task, though, can be relatively easily addressed by extensions to these frameworks, and we look forward to future work that might improve further on the results we showed.

\paragraph{Sequentially Controlled Generation}
Sequential control for text generation has been explored from many angles, from symbolic planning approaches \cite{meehan1976metanovel, lebowitz1987planning}, to keyword-based approaches \cite{yao2019plan} and concept, event and entity driven planning approaches \cite{rashkin2020plotmachines,peng2021guiding,alabdulkarim2021goal,han2022go}. We are the first, to our knowledge, to utilize a purely latent control structure based off of discourse structures. There is increasing interest in exploring how discourse can be used to guide generation \cite{ghazvininejad2021discourse,cohan-etal-2018-discourse}, from early works developing discourse schemas for generation \cite{mann1984discourse, stede1998dimlex} to evaluating creative generation pipelines \cite{hua-wang-2020-pair}. However, neither direction allows discourse structures to be explicitly controlled in generation.

\paragraph{Editing.}
Most existing neural models generate text in one-shot, from left to right. Recently, an emerging line of research~\cite{guu-etal-2018-generating,malmi-etal-2019-encode,kasner-dusek-2020-data} has explored editing as part of the generation pipeline to further improve the output quality, or satisfy certain desired constraints. Our work builds off of the MiCE framework~\cite{ross-etal-2021-explaining}, which was originally designed for generating contrastive explanations. Our observation that editing increased both the discourse adherence and the coherence of the generated text adds to a growing body of evidence that editing can play an important and modularized role in larger creative generation pipelines. We see editors such as simplification editors \cite{dong2019editnts}, factual correction editors \cite{cao2020factual} and stylistic editors \cite{kabbara2021post} each aimed at different desired attributes of text can possibly play a role in multi-attribute control \cite{li2022diffusion}.

Finally, we see overlaps as well to an earlier paradigm of generative modeling: Bayesian models for text like Latent Dirichlet Allocation (LDA) \cite{blei2003latent} and, more interestingly, sequential variants \cite{du2012sequential}. %
There is recent work marrying PPLM-style controlled text generation with topic modeling \cite{carbone2020etc}. Such directions might lead to more hierarchical, structural control. %

\section{Conclusion}

We have formalized a novel direction in controlled text generation: sequentially controlled text generation. We extended different techniques in controlled text generation to fit this direction, and have shown how a news discourse dataset can be used to produce news articles exhibiting human-like structure. We have explored what degrees of structural awareness yield the most human-like output: more structural control yields higher-quality output. And, we shown how to combine structural control with local editing. 
We have probed different parts of our pipeline to show the effects of each part.

\section{Acknowledgements}

Alexander Spangher acknowledges support from Bloomberg’s Data Science Ph.D. Fellowship. This work was conducted while Alexander was as intern at Bloomberg. We also acknowledge Jonathan May for helpful early conversations about this work. We acknowledge Vivienne Litzke for annotation effort and further conversations.

\section{Ethics Statement}

\subsection{Limitations}

A central limitation to our work is that the datasets we used to train our models are all in English. As mentioned previously, we used \newcite{choubey-etal-2020-discourse}'s \textit{NewsDiscourse} dataset, which consists of the sources: nytimes.com, reuters.com and xinhuanet.com. Although xinhuanet.com is a Chinese source, they used English-language articles. Additionally, we used an unlabeled news dataset from Kaggle\footnote{kaggle.com/snapcrack/all-the-news} for fine-tuning \texttt{GPT2-base} and for calculating some automatic metrics like \textbf{\% Unseen Words}. We filtered this dataset down to two English-languge, Western domains: nytimes.com and reuters.com in order to match the domains are closely as possible to the \textit{NewsDiscourse} dataset. 

Thus, we must view our work in discourse generation with the important caveat that non-Western news outlets may not follow the same discourse structures in writing their news articles. We are not aware of existing Van Dijk-style \cite{van2013news} datasets towards which we could provide an exact comparison. But, we hope in future work to look at other kinds of discourse structures that might exist in other languages.

\subsection{Risks}

There is a risk that the work will be used for misinformation or disinformation. This risk is acute in the news domain, where fake news outlets peddle false stories that attempt to \textit{look} true \cite{boyd2018characterizing, spangher2020characterizing}. Along this vein, there is the aforementioned work using discourse-structure to identify misinformation \cite{abbas2020politicizing, zhou2020fake}, and the risk in developing better discourse-aware generation tools is that these misinformation detectors might lose their effectiveness.

There is also a non-malicious misinformation risk, as large language models have been known to generate hallucinated information \cite{choubey2021mofe}. The more such threads of research are pursued \textit{without} an accompanying focus on factuality and truth, the more risk we run of polluting the information ecosystem. However, like others \cite{dathathri2019plug}, we see a value in continuing this direction of research, even if this current work is not the final output we wish to see being used by non-researchers in the world. It is one step along the way.

There is also a risk that news articles in either of our datasets contain potentially libelious or defamatory information that had been removed from the publishers' website after the dataset was collected. However, we do not release either of the datasets we use, so we do not see our actions as privacy-violating.

\subsection{Licensing}

Of the two datasets we used, \textit{NewsDiscourse} \cite{choubey-etal-2020-discourse} is published as a dataset resource in ACL 2020. They collected reuters.com and xinhua.net via crawling, and the nytimes.com from existing academically licensed datasets \cite{bhatia2015better, sandhaus2008new}. 

We were unable to ascertain the license for the Kaggle dataset. It has been widely used in the academic literature, including in papers published in ACL venues \cite{pathak2019breaking} and others \cite{alhuqail2021author}. We corresponded with the authors and opened a discussion question [URL withheld to preserve anonymity] seeking more information about the license. The authors are public about their desire to have their dataset used \footnote{https://components.one/datasets/all-the-news-2-news-articles-dataset/} and we have had independent lawyers at a major media company ascertain that this dataset was low risk for copyright infringement.

\subsection{Computational Resources}

The experiments in our paper required computational resources. We used 8 30GB NVIDIA GPUs, AWS storage and CPU capabilities. We designed all our models to run on 1 GPU, so they did not need to utilize model or data-parallelism. However, we still need to recognize that not all researchers have access to this type of equipment. We used Huggingface \texttt{GPT2-base} models for our predictive tasks, and will release the code of all the custom architectures that we constructed. Our models do not exceed 300 million parameters.

\subsection{Annotators}

We recruited annotators from professional networks. Both consented to annotate as part of the experiment in exchange for acknowledgement. One is a graduate student studying in Europe, and the other is a former journalist. One annotator is female, and the other is male. One is half-Asian and half-white identifying, the other is white. Both identify as cis-gender. This work passed IRB.

\bibliographystyle{acl_natbib}
\bibliography{custom}
\clearpage
\appendix

\section{Further Implementation Details}
\label{app:implementation_details}

\subsection{Discrimintor Implementation}
\label{app:implementation_details:discriminator}

\begin{figure}[t]
	\centering
    \includegraphics[width=.6\linewidth]{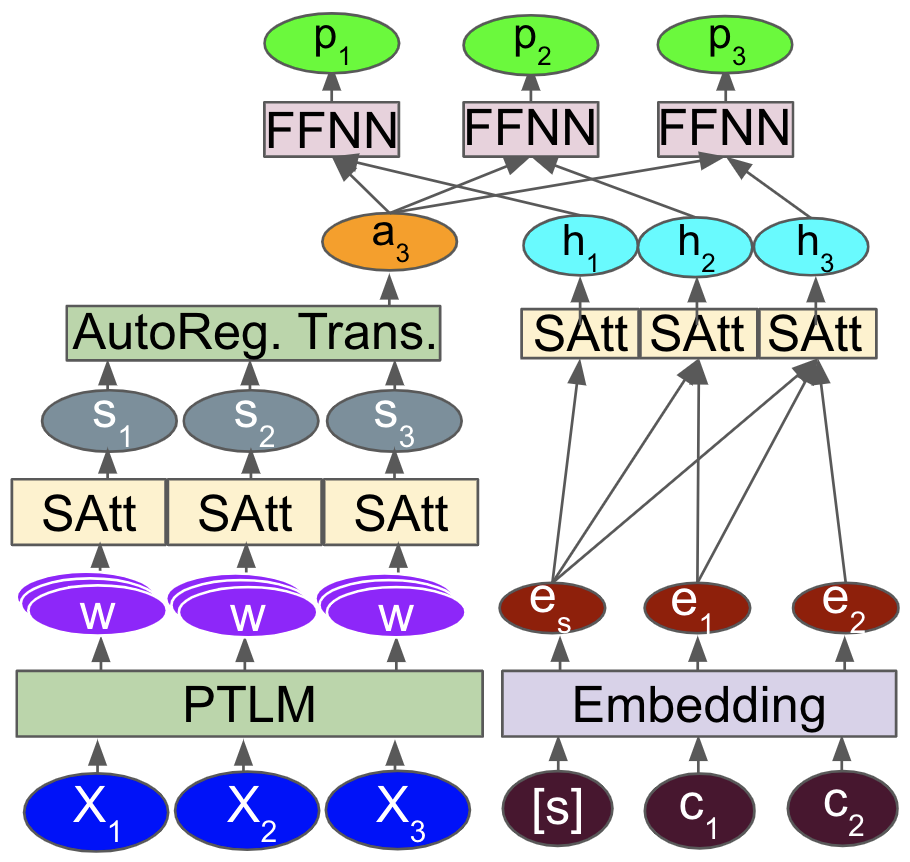}
	\caption{\textbf{Sentence classification model} for $k=3$ of a $3$ sentence document. Word embeddings ($\vec{w}_k$) for each sentence ($X_k$) are combined with self attention ($s_k$). A transformer contextualizes $s_k$ ($a_k$) with $s_{<k}$. Labels $\vec{c}$ are embedded ($e$) and self-attention generates label vectors ($h_k$). $a_k$, $h_k$ are combined for predictions ($\vec{p}$).}
	\label{fig:nn_model:diagram}
\end{figure}

We tested 122 different discriminator variations. A summary of the major architectural iterations is shown in Table \ref{tab:discriminator_f1}. We describe each variation as follows; the top-performing variation, with a subset of input sentences and labels, is shown in Figure \ref{fig:nn_model:diagram}.

Contextualized word vectors ($\vec{w}$) from a PTLM (we experimented with either GPT-2 or a RoBERTa as in \newcite{spangher2021multitask}) are obtained for each sentence, and are combined using self-attention. Switching to GPT-2 yielded a 16-point F1-macro score drop. Hidden-State Control, based on \cite{dathathri2019plug}, relies on perturbations to the state variable H from the naive language model to generate word-probabilities $p(x_i | X_{<k}, x_{<i}, \vec{c}) = p(x_i | H, \vec{c}) = p(\vec{c} | H, x_i)p(x_i | H)$. So, we need to use the same PTLM for the language model as we do for the discriminator. %
We do \textit{not} have the same restriction on Direct Probability Control \cite{yang2021fudge}, as the probabilities are directly multiplied and thus do not need to share any architectural components. For the sake of an apples-to-apples comparison on the mechanism of control, though, we use a GPT-2 model for the PTLM layer in our discriminator.

Next, we tested either embedding each sentence separately in batch, or embedding the entire document (+Flattened Sentences). Embedding the entire document yielded a 3 point F1-macro increase.

These sentence vectors are then contextualized: we tested an LSTM layer (+LSTM) to contextualize these vectors and an autoregressive transformer layer (+Transformer)\footnote{With 2 layers and 2 attention heads}. Using transformer yielded a 6 point F1-macro increase. We next fine-tuned the GPT-2 LM using it's LM head on an unlabeled, 30K article news corpus. This yielded a 3 point F1-macro increase.

To incorporate label information as input to the model (as in the Past and Full variants) we embed each label using a learned embeddings layer, and then we combine these embeddings using self-attention\footnote{This architecture allows us to capture structural dependencies between labels better than approaches like a CRF layer, which cannot easily be extended beyond linear-chain operations.}. Experimenting with a different window size yielded a 5 point F1-macro increase. We find that a window of $3$ yields the best-performing discriminator.

Finally, a feed forward classifier combines the sentence vector with the label vector. We find that sharing the PTLM improves accuracy, but not other layers. %

\begin{table}[t]
    \centering
    \small
    \begin{tabular}{lr}
    \toprule
    Discriminator Version &     F1 Macro \\
    \midrule
    RoBERTa Baseline         &  0.62 \\
    \midrule
    GPT-2 & \\
    + Contextualizing Layer & \\
    \quad LSTM       &  0.46 \\
    \quad Transformer    &  0.52 \\
    + Flattened Sentences    &  0.55 \\
    + LM Fine-Tuned with News Corpus  &  0.58 \\
    + Labels & \\
    \quad Full     &  0.58 \\
    \quad Window=7 &  0.61 \\
    \quad Window=5 &  0.62 \\
    \quad Window=3 &  \textbf{0.63} \\
    \quad Window=2 &  0.62 \\
    \bottomrule
    \end{tabular}
    \caption{F1 Macro on main prediction head, $p_k$, for different discriminator variations. RoBERTa baseline is from \newcite{spangher2021multitask}. GPT-2 variations described in body.}
    \label{tab:discriminator_f1}
\end{table}

\begin{figure}[t]
     \centering
     \subfloat[]{
        \includegraphics[width=.48\linewidth]{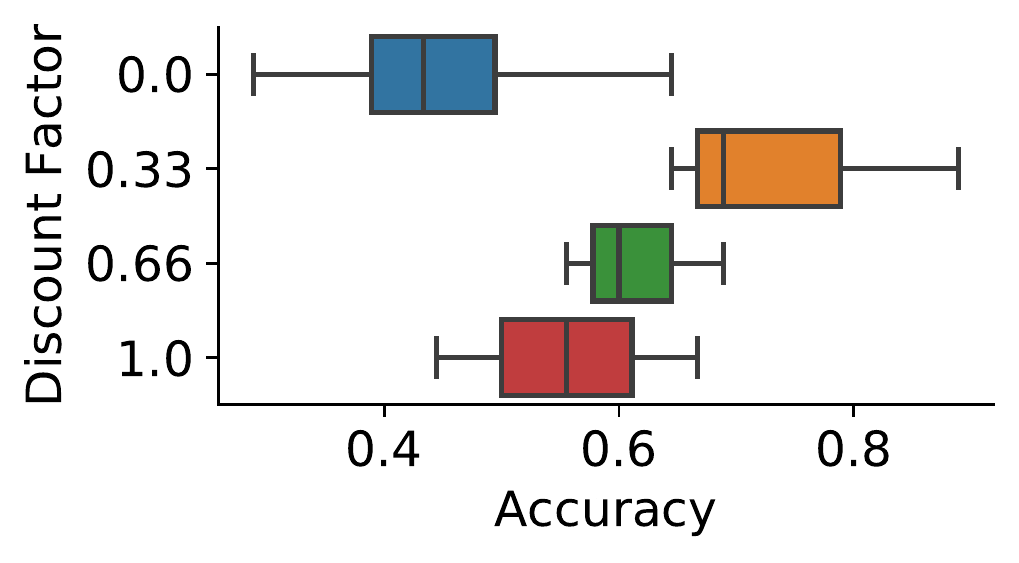}
        \label{fig:boxplots:exp:acc}
     }
    \subfloat[]{
        \includegraphics[width=.48\linewidth]{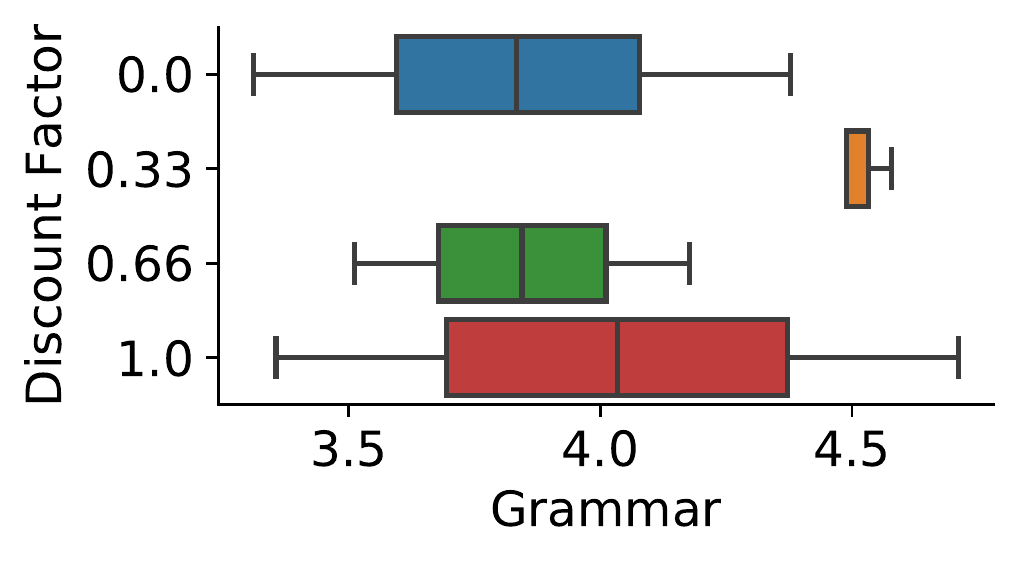}
        \label{fig:boxplots:exp:grammar}
    } \\ 
    \subfloat[]{
        \includegraphics[width=.48\linewidth]{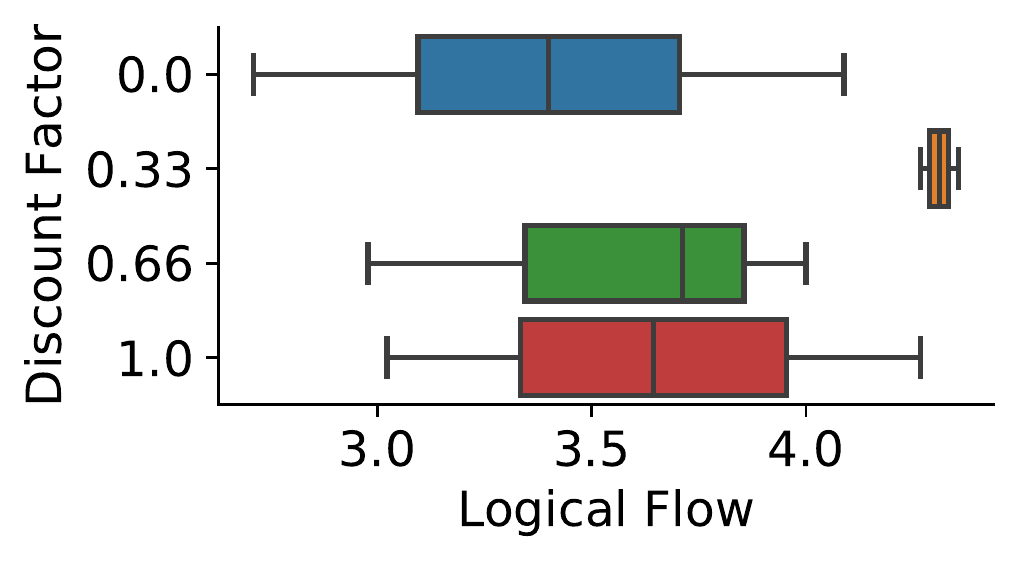}
        \label{fig:boxplots:exp:coherence}
    }
    \subfloat[]{
        \includegraphics[width=.48\linewidth]{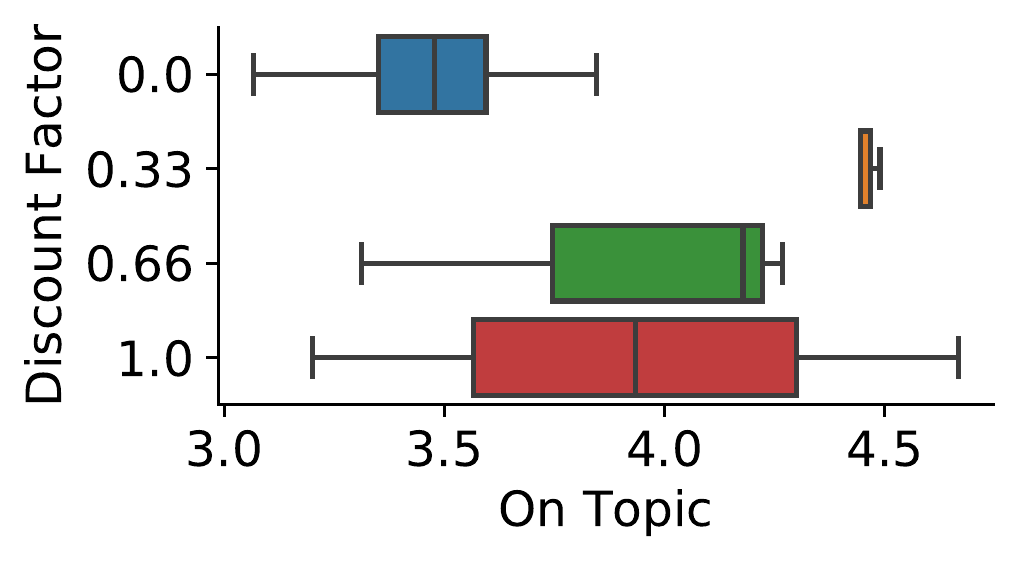}
        \label{fig:boxplots:exp:topical}
    }
    \caption{The effect of Discount Factor $b$, across different pipelines and hyperparameters.}
    \label{fig:boxplots:exp}
\end{figure}

\subsection{Details on Hyperparameters}
\label{app:hyperparam}
\subsubsection{Discount Factor, $b$}
To impose further structural control, we impose a prior on $t_3$ that acts as a discount factor. In words, we downweight the discriminator probabilities for control codes that are farther away from the current sentence being generated. The form of our prior is: $t_3 = \prod_{j=1}^{S} m(i, j) p(c_{j} |x_i, x_{<i}, X_{<k}, c_{<j})$, where $m(i, j) = b^{|i-j|}$. We experiment with $b=[.33, .66, 1]$. So, the lower the discount factor, $b$, the more the current, local control code matters. When $b=0$, the \textit{local-only} variant of our discriminator, Equation \ref{eq:local_only}, is expressed by default.

We see in Figure \ref{fig:boxplots:exp} that discount factor $b$ has a non-linear effect on the output. In accordance with our prior results, $b=0$ is the lowest-performing variant across all four human-quality metrics. $b=.33$ seems to be the most effective discount factor overall, and yields the best output for accuracy and logical flow, while $b=1$ yield the best-performing output for grammar and topicality. We conclude that a finer-grained balance of local control and structural control might be important overall, but in some cases more structural control might help as noted previously.

\subsubsection{Hidden-State Control (HS)}
\label{app:hyperparam:pplm}

In \newcite{dathathri2019plug}, authors find anywhere between $3$ and $10$ backpropagation steps is acceptable. In this work, we use $10$ steps with a small step size. We also test different regularizations, also explored in \cite{dathathri2019plug}, on the output logits generated from $\hat{H}$. We experiment with different hyperparameters for one of the regularizations: $\hat{l}=\gamma \hat{l} + (1-\gamma) l^0$ where $l^0$ is the naive, unperturbed logits. We experiment with different values of $\gamma$ from $0$ (fully unperturbed) to $1$ (fully perturbed).

\subsubsection{Direct-Probability Control (DPC)}
\label{app:hyperparam:fudge}

Authors in \cite{yang2021fudge} offer an innovation by training their classifier $p(c|x)$ to consider subsequences $p(c|x_1,...x_i)$ for all $i$, ostensibly improving the accuracy of their joint probability calculation while midsequence. This is in contrast to \newcite{dathathri2019plug}'s training regimine, which only considers full sequences $p(c|x_1,...x_n)$. However, \newcite{yang2021fudge} do not provide ablations to show whether it is this training regimine, or their direct calculation of $p(x)p(c|x)$, which is responsible for the improvements they observe. In this work, we perform this ablation and find that it has negligible difference, according to automatic evaluation metrics. We also introduce a mean fusion \cite{stahlbergsimple} into the $p(x)p(c|x)$ joint likelihood:  $\gamma p(c|x) + (1-\gamma)p(x)$ and test different values of $\gamma$.

\section{Automatic Metrics List}
\label{app:metrics}

Here, we discuss the automated metrics reported in Table \ref{tab:results}. They are largely based off metrics proposed in \newcite{see2019massively}.

\subsection{Metrics Reported in Paper}

\paragraph{Label Probability}: We measure the label probability assigned to the gold-truth class label given in our input sequence: $p(c | c_{<s}, x_i, x_{<i}, X_{<s})$. We use head $p$, or the current head, in the discriminator shown in Figure \ref{fig:nn_model:diagram}.

\paragraph{Perplexity}: Perplexity is calculated using the fine-tuned GPT-2 model, which we fine-tuned on $30,000$ news articles.

\paragraph{Diverse N-grams}: We measure the likelihood that an n-gram in one sentence will be unique compared with the entire document. In other words:
\begin{multline}
\text{Diverse N-Grams}(s, d) =\\ \frac{\text{\# unique n-grams in sentence } s}{\text{\# n-grams in document } d}
\end{multline}
We calculate the set of n-grams per document as the total number of 1,2,3-grams in that document. We calculate one measurement per sentence in the document, and average these scores together.%

\paragraph{Sentence Length}: We measure the total number of words in the sentence, based on word-level tokenization using \texttt{https://spacy.io/}.%

\paragraph{Unseen Words}: We use an external corpus of $30,000$ news articles to determine a typical, large news vocabulary. Any words that are outside of this vocabulary are considered ``Unseen Words''. For our purposes, we are most interested in exploring malformed words, which are sometimes generated by the language model. However, unseen words might also be proper nouns.

\section{Generation-Baseline \#1: Prompting. Further Details}
\label{app:baseline_2}

As a baseline, we train a language model to directly calculate $p(x_i | x_{<i}, X_{<s}, \vec{c})$, following \cite{keskar2019ctrl}. We design the following prompt structure to simulate baseline, past-aware and full-sequence control variants.

\noindent\textbf{Baseline}:

\texttt{Headline: <Headline> Labels: <Current Label> Sentences: <Sentence 1> <Sentence 2>... <Sentence s>}

\noindent\textbf{Past-Aware}:

\texttt{Headline: <Headline> Labels: <Label 1>, <Label 2> ... <Label k> Sentences: <Sentence 1> <Sentence 2>... <Sentence s>}

\noindent\textbf{Full-Sequence}:

\texttt{Headline: <Headline> Labels: <Label 1>, <Label 2> ... <Label s> Current Position: <i> Sentences: <Sentence 1> <Sentence 2>... <Sentence s>}

The prompts are specific to current sentence being generated. We first start by generating sentence $1$, whereby the prompt for \textbf{Baseline} and \textbf{Past-Aware} is both:

\texttt{Headline: <Headline> Labels: <Label 1> Sentences:}

Then, we let the model generate the first sentence and stop when we generate the $<EOS>$ character. We then regenerate the prompt to include the previously generated sentence and update the tags, so \textbf{Baseline} becomes:

\texttt{Headline: <Headline> Labels: <Label 2> Sentences: <Sentence 1>}

and \textbf{Past-Aware} becomes:

\texttt{Headline: <Headline> Labels: <Label 1> <Label 2> Sentences: <Sentence 1>}

We continue in this fashion, resetting the prompt each time, until we have finished generating sentences for all the tags in our input data.

The Full-Sequence process is very similar, except we do not need to update the label-space, since by default the model is exposed to the full sequence of tags before generation.

\section{Editing}
\label{app:editing}

In this section, we describe the various components of the editing model. First, we note the differences in our approach and \newcite{ross-etal-2021-explaining}'s method. Then, we discuss the infilling model and the discriminator.

\begin{table*}[t]
	\centering
    \small
    \begin{tabular}{p{2cm}p{6.5cm}p{6.5cm}}
    \toprule
    Discourse Tag &  Pre-editing &  Post-editing \\
    \midrule
        \hlConsequence{Consequence} &  
            The company \hlpink{has} \hlpink{already} \hlpink{spent} \$ 23 billion in Medicare, seeking antitrust clearance. & 
            The company \hlgreen{also} \hlgreen{plans} \hlgreen{to} \hlgreen{buy} \$ 23 billion in Medicare, seeking antitrust clearance. \\
        \hline
        \hlExpectation{Expectation} &
            Volvo Car \hlpink{dropped} in the first quarter after a trade row over Chinese car makers. & 
            Volvo Car \hlgreen{is} \hlgreen{expected} \hlgreen{to} \hlgreen{close} \hlgreen{lower} in the first quarter after a trade row over Chinese car makers. \\
        \hline
        \hlEvaluation{Evaluation} &
            The deal \hlpink{values} Wind Energy, which has operations \hlpink{offshore} in New York. &
            The deal \hlgreen{is} \hlgreen{significant} \hlgreen{for} Wind Energy, which has operations \hlgreen{mostly} in New York. \\
        \hline
        \hlCurrentContext{Current Context} &
            8 billion shares \hlpink{sold} in all of \hlpink{2015}. &
            8 billion shares \hlgreen{were} \hlgreen{traded} in all of \hlgreen{China}. \\
        \hline
        \hlExpectation{Expectation} &
            The deal \hlpink{comes} as insurers and drugmakers struggle with competition \hlpink{from} Medicare prescription drugs. &
            The deal \hlgreen{could} \hlgreen{stall} as insurers and drugmakers struggle with competition \hlgreen{for} Medicare prescription drugs. \\
    \bottomrule
    \end{tabular}
	\caption{A selection of sentences and the edit operations performed on them. The editor focuses on (a) temporal relations, (b) conditional statements (c) explicit discourse markers (e.g. ``expect'') and correct grammar.}
\end{table*}

\subsection{Key Differences}
\label{app:editing:comparison}

\newcite{ross-etal-2021-explaining} designed their editor to flip classifier predictions. So, they edited input $x \rightarrow \hat{x}$ until $\argmax_c p(\vec{c} | \hat{x}) \neq \argmax_c p(\vec{c} | x)$. Then, $\Delta(x, \hat{x})$ was given as the explanation for the flip. We are not concerned with flipping predictions so much as maximizing the probability of the ground truth label. So, we design our objective to be $x \rightarrow \hat{x}$ until $p(c | \hat{x}) > p(c | x)$.

To understand why the loss-gradient on the input can provide feature importance, consider the first-order Taylor approximation of the loss, $l(x) \approx l(a) + l'(a)(x-a)$. Here, the gradient of the loss at $a$, $l'(a)$, can be seen as a set of linear weights similar to logistic regression coefficients, which are commonly used for feature importance. 

We also wished to restrict editing to \textit{explicit discourse markers}, spuriously correlated words, so we heuristically excluded all Proper Nouns, Named Entities (except DATE) or adjectives from the edit candidate set. Table \ref{tab:nd_features} shows explicit discourse markers in the news discourse context. Here, we show the top words associated with each discourse class\footnote{Most positive coefficients of a Logistic Regression Classifier that takes as input a sentence and predicts it's discourse class}. Some words effect the tense of the sentence\footnote{Top verbs in \textit{Expectation} are almost all present-tense, while top verbs in \textit{Previous Event} are almost all past-tense}, others inject epistemological uncertainty\footnote{Top verbs in \textit{Evaluation} are all ``say'' verbs, while verbs in \textit{Current Context} are based on observable events}, still others time-peg events to certain days\footnote{Top \textit{Main Event} nouns are nearly all weekday names}.

\subsection{Infilling Model}

We train a label-aware infilling model in a similar method as \newcite{ross-etal-2021-explaining}. Our prompt is:

\texttt{label: <label> text: Lorem Ipsum <mask> Lorem <mask> Ipsum.}

Where the masks replace high-salience words, which we discovered as described above. We format samples using sentences in our training dataset, and train a T5 model as described by the authors. 

\subsection{Possible Improvements}

We note that this infilling method directly models $p(\hat{x} | M(x), c)$, i.e., the likelihood of infilled words given a label and a masked sentence. Another possible approach to this problem would be to use a naive infiller and Bayes rule as done in the generation phase of this paper to generate logits $p(\hat{x} | M(x)) p(c | \hat{x}, M(x))$. This could possibly improve the editor for the same reasons \newcite{dathathri2019plug} and \newcite{yang2021fudge} observed an improvement over CTRL \cite{keskar2019ctrl}. %

Another aspect of the editor that we noticed was that it could sometimes degrade the coherency and topicality of the document. This is especially evident in the \textbf{Human} trials. We partially addressed this by selecting candidate edits based off the perplexity of the whole document. We could have mitigated this further by giving our infiller the entire document as context \footnote{I.e. we could have trained a model based on $p(\hat{x} | M(x), X_{<s}, c)$, instead of $p(\hat{x} | M(x), c)$}. %

\section{Further Methods Comparison}
\label{app:method_comparison}

The standard controlled text generation setup is typically expressed as follows:
\begin{equation}
\label{eq:standard_controlled_gen}
p(x|c) = \prod_{i=1}^n p(x_i | x{< i}, c)
\end{equation}
where $x$ is the output sequence and $c$ is a single control code (for example: sentiment \cite{dathathri2019plug}). Here, $x$ is a single sentence (or paragraph) of $n$ words, factorized autoregressively into words $x_i$ and previous words $x_{<i}$.

Previous approaches to controlled text generation \cite{dathathri2019plug, yang2021fudge} factorize the right term of Equation \ref{eq:standard_controlled_gen} as follows:
\begin{equation}
	\label{eq:factorized_standard_control}
	p(x_i | x_{<i}, c) \propto p(x_i | x_{<i})p(c|x_i, x_{<i})
\end{equation}
As in Equation \ref{eq:standard_controlled_gen}, this factorization decomposes  our sequentially controlled text generation model into an uncontrolled language model and a control-code model. The key difference between Equation \ref{eq:factorized_standard_control} and \ref{eq:factorized_sequential_control} is in the second term, i.e. how we choose to model the control codes (the difference in the first term is simply a rather trivial extension of a naive language from a sentence-to-paragraph generation to a document-generation context). 

We show a direct comparison of all of our generation approaches in Figure \ref{fig:boxplots:discrim}. Here, we show %
that Direct Probability Control has the best effect over Naive GPT-2 for class-accuracy and, surprisingly, perhaps, Grammar and Topicality as well.

\section{Ovid's Unicorn Is Not Structural}
\label{app:ovids_unicorn}

We annotate of the famous Ovid's Unicorn news article generated and presented by the original GPT-2 authors \cite{radford2019language}. 

We analyse this article as we have analyzed our generation models Section \ref{sct:results}. One of our annotators gave each sentence the Van Dijk discourse label that best fits \cite{van2013news}, and the other assessed whether it actually fit. This is not an apples-to-apples comparison with the \textbf{Label Acc.} column in Table \ref{tab:results}, because we are assessing the accuracy of the label that \textit{we} chose \textit{after} reading the text.

\begin{figure}[t]
    \centering
    \includegraphics[width=.8\linewidth]{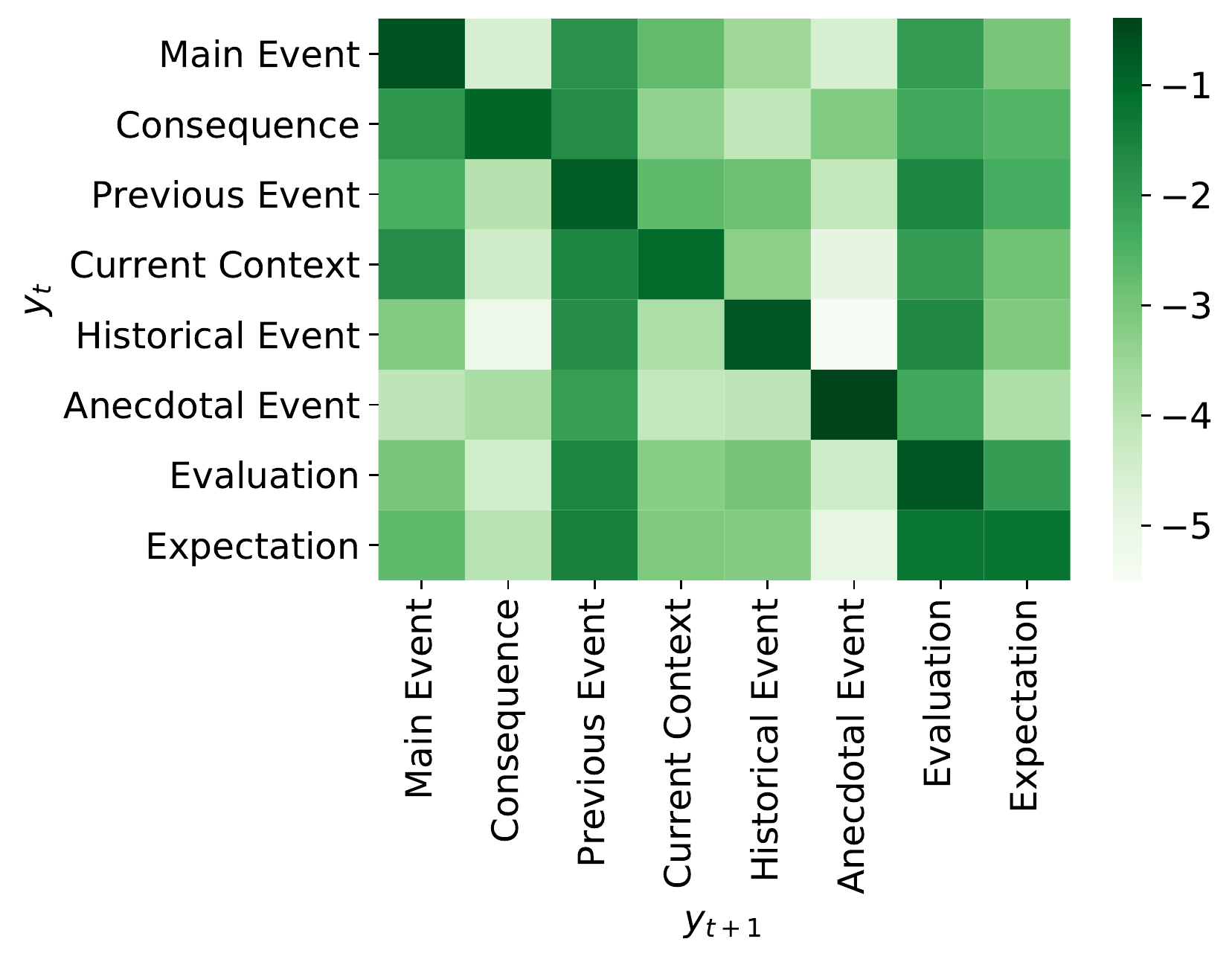}
    \caption{Transition Probability Matrix (log likelihood) for tag sequences.}
    \label{fig:transition}
\end{figure}

\begin{figure}[t]
     \centering
     \subfloat[]{
        \includegraphics[width=.48\linewidth]{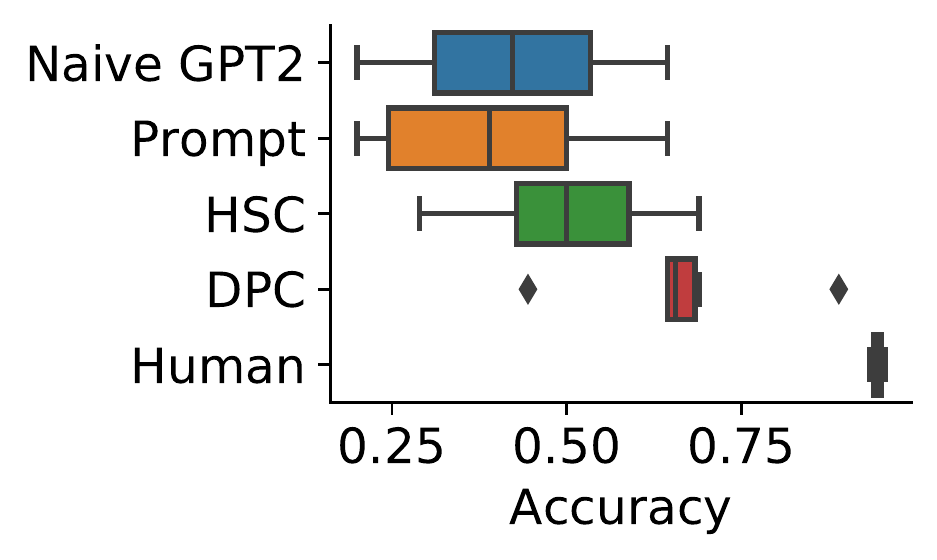}
        \label{fig:boxplots:generation:acc}
     }
    \subfloat[]{
        \includegraphics[width=.48\linewidth]{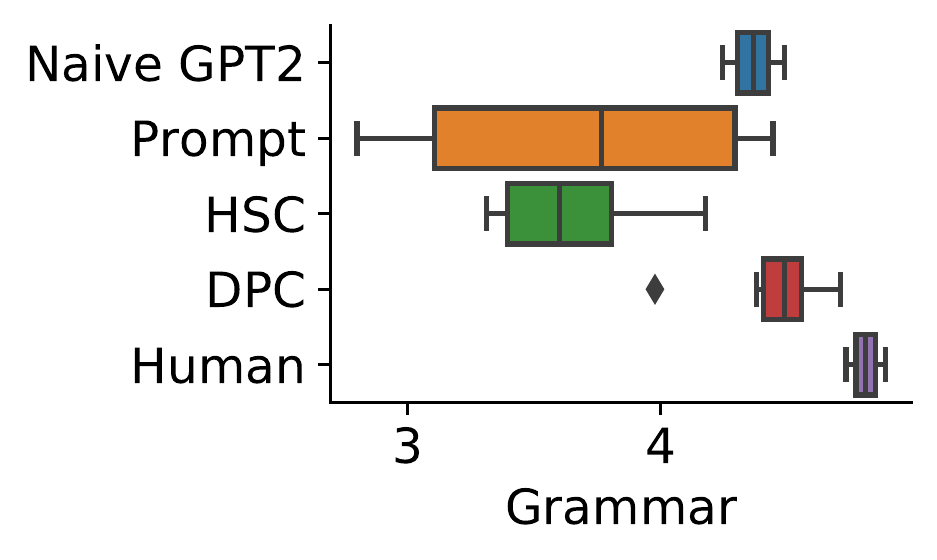}
        \label{fig:boxplots:generation:grammar}
    } \\ 
    \subfloat[]{
        \includegraphics[width=.48\linewidth]{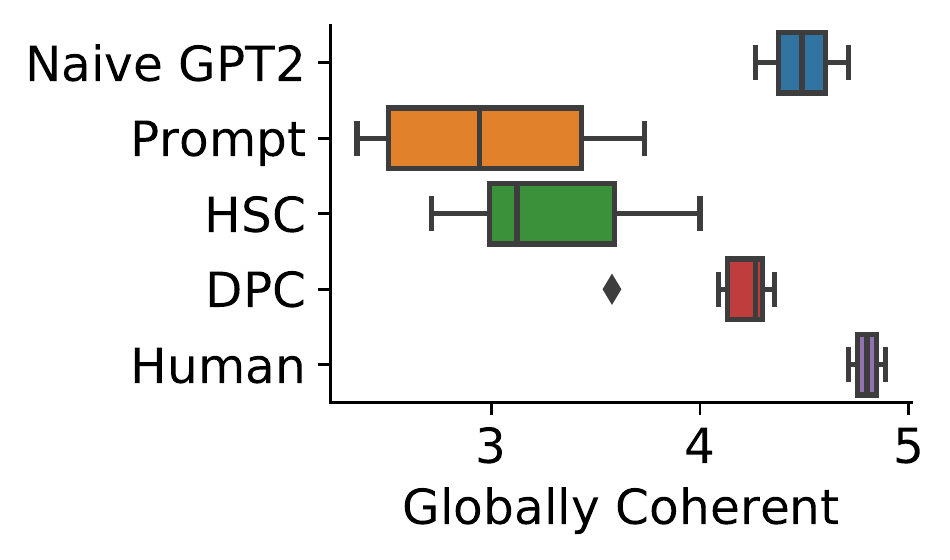}
        \label{fig:boxplots:generation:coherence}
    }
    \subfloat[]{
        \includegraphics[width=.48\linewidth]{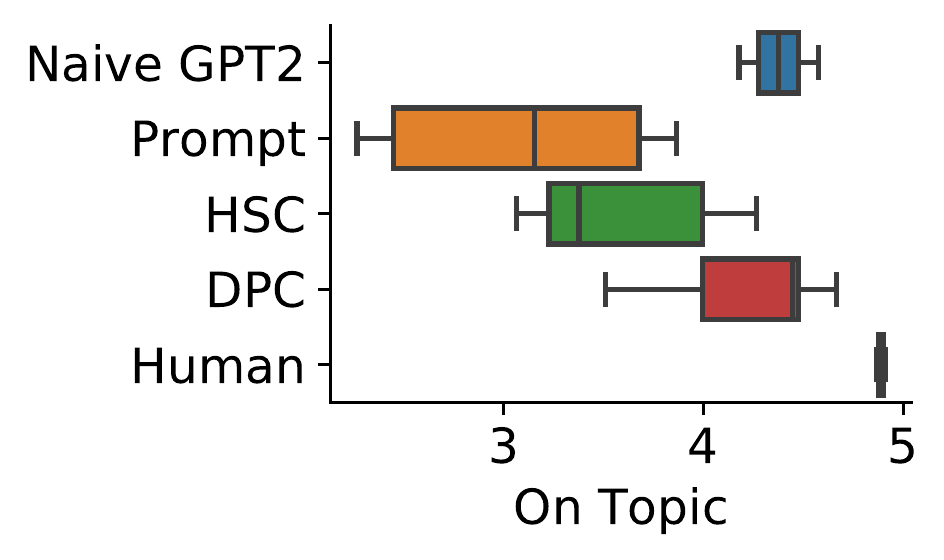}
        \label{fig:boxplots:generation:topical}
    }
    \caption{Different generation methods, across different pipelines and hyperparameters.}
    \label{fig:boxplots:generation}
\end{figure}

\begin{table}[t]
\centering
\small
\begin{tabular}{lr}
\toprule
Article Source &     Average Log-Likelihood \\
\midrule
Test Set (5/50/95 Percentile)  &  -1.28/-1.60/-2.01\\
Ovid Unicorn's & -2.24 \\
\bottomrule
\end{tabular}
\caption{Log-Likelihood of Tag-Sequence, according to simple bi-gram model $p(c_{t+1}| c_t)$, trained by counting tag sequences in the training dataset. 5th/50th/95th percentiles shown for test set.}
\label{tab:ll_comp}
\end{table}

We next measured the likelihood that an article with the discourse structure of Ovid's Unicorn would exist naturally. We build a simple bigram model for tags, $p(c_{t+1} | c)$, to calculate the total probability of a tag sequence.%
We show in Figure \ref{fig:transition}, the typical transitions between discourse labels in the news discourse dataset. We fit our simple bigram model using label sequences in the training dataset, and calculate average log-likelihood of the tag sequence for each document in our test dataset. The median of across these is shown in Table \ref{tab:ll_comp}. As can be seen, sequences in the test dataset are far more likely than the Ovid's unicorn article, which falls outside of the 95th percentile of the distribution of typical articles.

\begin{table}[t]
\small
\centering
\begin{tabular}{p{2.2cm}lll}
\toprule
Discourse Label  & \multicolumn{2}{l}{Top words} \\
\midrule
Main Event       &    monday & cooperation &       shot \\
Consequence      &    closed &      showed &     issued \\
Previous Event   &   comment &    declined &        agency \\
Current Context  &      shot &       prime &        groups \\
Historical Event &      2015 &        2016 &       2017 \\
Anecdotal Event  &      want &        told &        old \\
Evaluation       &       say &       think &       told \\
Expectation      &  expected &      likely &   continue \\
\bottomrule
\end{tabular}
\caption{Top predictive words for each discourse type (top positive $\beta$ coefficients for a Logistic Regression trained to predict $y=$ news discourse tag per sentence using and $X=$ a bag of words representation of each sentence).}
\label{tab:nd_features}
\end{table}

\subsection{Van Discourse-based Schema Introduced in \newcite{choubey-etal-2020-discourse}}
\label{app:van_dijk_schema}

The schema used for \textit{News Discourse}, introduced by \cite{choubey-etal-2020-discourse}, was based off the schema introduced by \newcite{van2013news}. As such, the classification guidelines were:

\paragraph{\hlMainEvent{Main Event}}: The major subject of the news report. It can be the most recent event that gave rise to the news report, or, in the case of an analytical news report, it can be a general phenomenon, a projected event, or a subject.

\paragraph{\hlConsequence{Consequence}}: An event or phenomenon that is caused by the main event or that directly succeeds the main event.

\paragraph{\hlPreviousEvent{Previous Event}}: A specific event that occurred shortly before the main event. It either directly caused the main event, or provides context and understanding for the main event.

\paragraph{\hlCurrentContext{Current Context}}: The general context or world-state immediately preceding the main event, to help the readers better understand and contextualize the main event. Similar to \textbf{Previous Event}, but not necessarily tied to  a specific event.

\paragraph{\hlHistoricalEvent{Historical Event}}: An event occurring more than 2 weeks prior to the main event. Might still impact or cause the main event, but is more distal.

\paragraph{\hlExpectation{Expectation}}: An analytical insight into future consequences or projections made by the journalist.

\paragraph{\hlEvaluation{Evaluation}}: A summary, opinion or comment made by the journalist on any of the other discourse components.

\paragraph{\hlAnecdotalEvent{Anecdotal Event}}: Sentences describing events that are anecdotal, such events may happen before or after main events. Anecdotal events are specific events with specific participants. They may be uncertain and can't be verified. A primary purpose of this discourse role is to provide more emotional resonance to the main event.

In Table \ref{tab:nd_features} we attempt to provide more insight into different News Discourse elements by modeling using Logistic Regression. %

\section{Annotation}
\label{app:annotation}

\begin{figure*}[t]
    \centering
    \includegraphics[width=.6\linewidth]{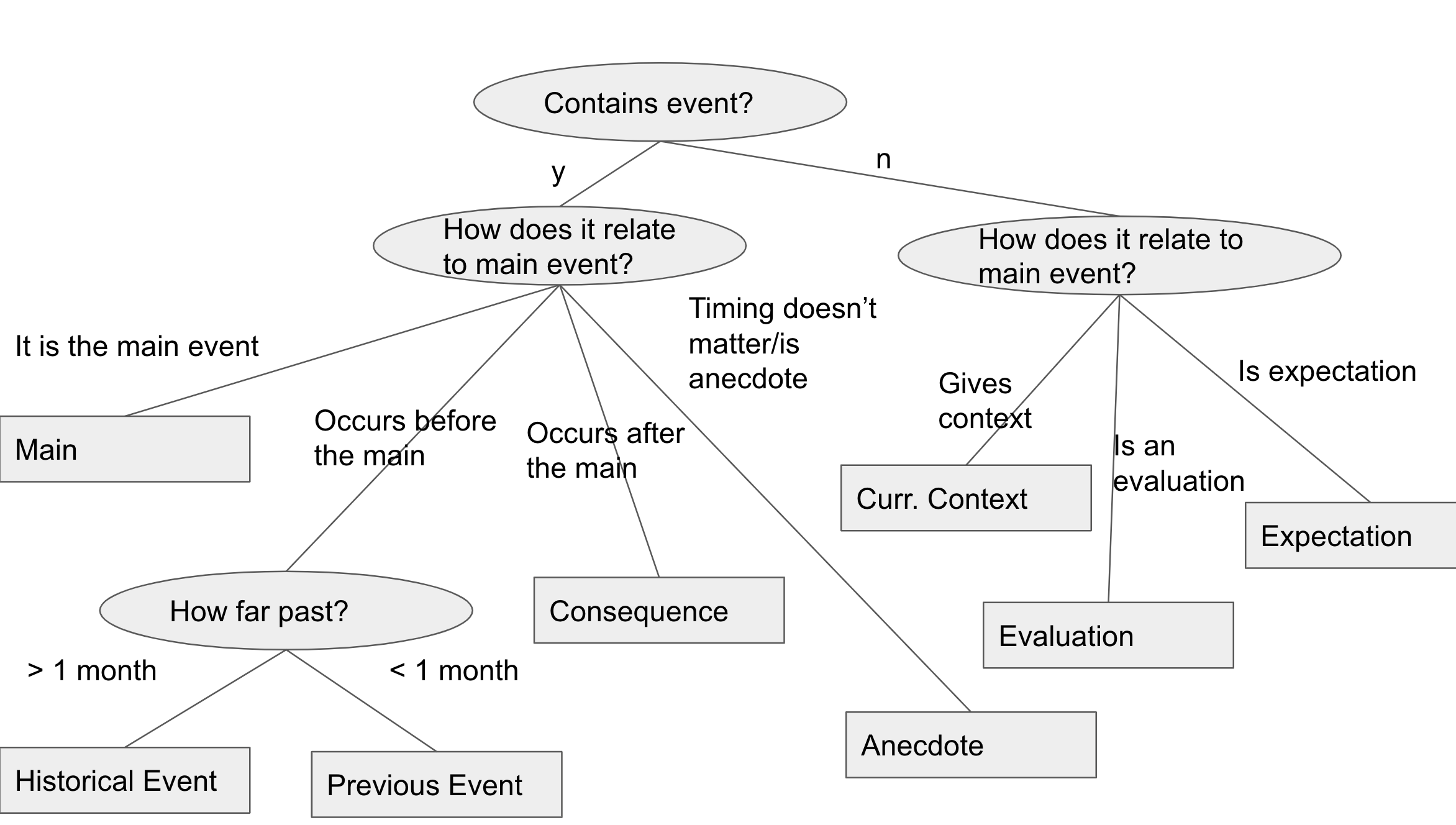}
    \caption{Tree shown to annotators for reference on manual annotation task.}
    \label{fig:decision_tree}
\end{figure*}

\begin{figure*}[t]
    \centering
    \includegraphics[width=\linewidth]{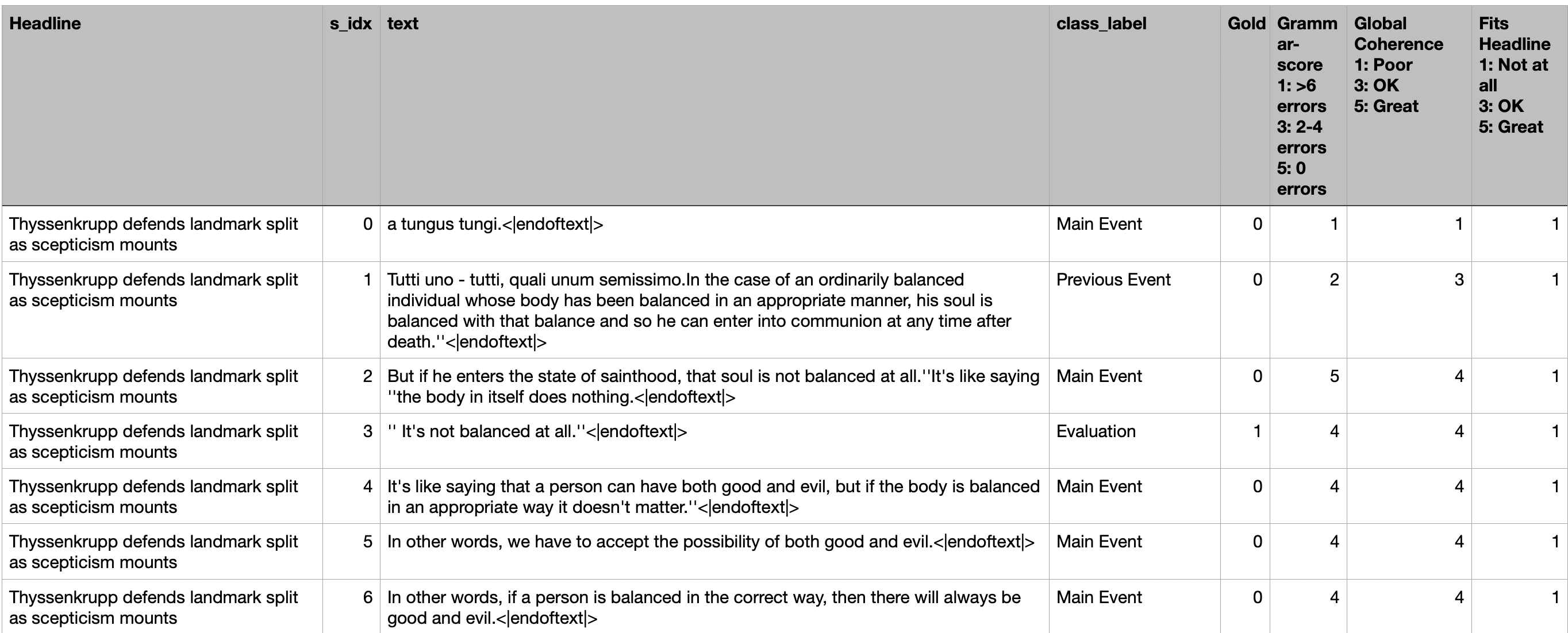}
    \caption{Visual of the annotation task interface that we asked our annotators to use. We presented annotators with class labels and asked them to simply determine Y/N whether the label was accurate. We also added a question to probe topicality. %
    (\textbf{Prompting Baseline} is the method generating the text currently seen in the interface.)}
    \label{fig:annotation_interface_2}
\end{figure*}

We recruit two manual annotators, one with $>1$ year and the other with $>4$ years of journalism experience. Both annotators offered to perform these tasks voluntarily in exchange for acknowledgement.

For their reference, we showed the annotators the label definitions (shown in Section \ref{app:van_dijk_schema}) and a decision-tree (shown in Figure \ref{fig:decision_tree}). The decision-tree breaks down key components of discourse reasoning.%

Additionally, we gave them training annotation questions for practice. %
For the training task, they were asked to view human-written sentences from 10 articles and go through the step-by-step question process based on the decision tree. These labels were checked with the gold labels from the training dataset, and they trained until they were answering questions with >80\% accuracy.

The interface we used to collect annotations is shown in Figure \ref{fig:annotation_interface_2}. Annotators were blind to the method that generated the text but were shown the desired true labels and simply had agree Y/N if the label fit\footnote{An earlier interface that asked annotators to assign their own tags was too difficult.}

For Grammar, we asked them to count the number of grammar mistakes per sentence \texttt{(1: >6, 3:2-4, 5:0)}. For Logical Flow, we used a qualititative metric \texttt{(1: ``Poor'', 3: ``OK'', 5: ``Great'')}. For Topicality, we also used a qualitative metric \texttt{(1: ``Not at all'', 3: ``OK'', 5: ``Great'')}

\end{document}